\documentclass[conference]{IEEEtran}
\usepackage[utf8]{inputenc}
\usepackage{multicol,url,color}
\usepackage{multirow}
\usepackage[table]{xcolor}
\usepackage{amsmath,amssymb,amsfonts}
\usepackage{algorithm}
\usepackage[noend]{algpseudocode}
\usepackage{graphicx}
\usepackage[multiple]{footmisc}
\usepackage{caption}
\usepackage{comment}
\usepackage{textcomp}
\usepackage{amssymb}
\usepackage{hyperref}
\usepackage{booktabs}
\usepackage[font=small,labelfont=bf]{caption}

\usepackage{flushend}

\makeatother
\usepackage{float}
\usepackage{mathtools} 
\usepackage{graphics}
\usepackage{placeins}
\usepackage{array}
\usepackage{enumitem}
\usepackage{cite}
\usepackage{color}
\usepackage{listings}
\usepackage{tikz}

\newcommand{\vect}[1]{\mathbf{#1}}

\newcommand{\note}[1]{\textcolor{blue}{NOTE #1}}

\newcommand{\Wskip}[1]{ }
\definecolor{green}{rgb}{0.0, 0.8, 0.0}

\newcommand{\nnz}{\textit{nnz}}

\title{\huge{FusedMM: A Unified SDDMM-SpMM Kernel for \\ Graph Embedding and Graph Neural Networks}}

\author{\IEEEauthorblockN{
    Md. Khaledur Rahman,
    Majedul Haque Sujon,
    and
    Ariful Azad} Indiana University, Bloomington, IN, USA
}
\author{\IEEEauthorblockN{Md. Khaledur Rahman, Majedul Haque Sujon, and Ariful Azad}
\IEEEauthorblockA{\textit{Luddy School of Informatics, Computing, and Engineering} \\
\textit{Indiana University Bloomington},
IN, USA \\
Email: \{morahma, msujon, azad\}@iu.edu}
}

\date{}

\begin{document}
\thispagestyle{plain}
\pagestyle{plain}
\setcounter{page}{1}
\renewcommand\citepunct{, }
\maketitle

\begin{abstract}
    We develop a fused matrix multiplication kernel that unifies sampled dense-dense matrix multiplication and sparse-dense matrix multiplication under a single operation called FusedMM. By using user-defined functions, FusedMM can capture almost all computational patterns needed by popular graph embedding and GNN approaches. 
    
    FusedMM is an order of magnitude faster than its equivalent kernels in Deep Graph Library.
    The superior performance of FusedMM comes from the low-level vectorized kernels, a suitable load balancing scheme and an efficient utilization of the memory bandwidth. 
    FusedMM can tune its performance using a code generator and perform equally well on Intel, AMD and ARM processors.
    FusedMM speeds up an end-to-end graph embedding algorithm by up to $28\times$ on different processors. The source code is available at \url{https://github.com/HipGraph/FusedMM}.

\end{abstract}


\section{Introduction}
\begin{figure*}[!t]
    \centering
    \includegraphics[width=\linewidth]{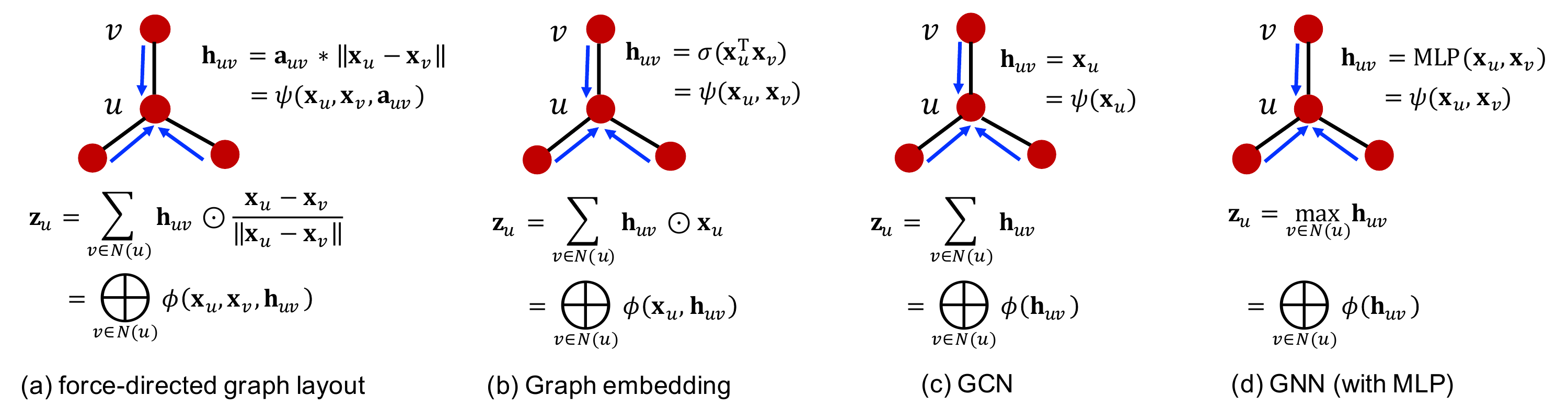}
    \vspace{-10pt}
    \caption{Message passing models in (a) force-directed graph drawing, (b) graph embedding (e.g., in VERSE~\cite{tsitsulin2018verse}), (c) graph convolutional network (GCN), and (d) general graph neural networks with multilayer perceptron (MLP). In all cases, the message $\vect{h}_{uv}$ that passes from $v$ to $u$ is a function of the form $\psi(\vect{x}_u, \vect{x}_v, \vect{a}_{uv})$, where $\vect{x}_u$ and $\vect{x}_v$ are node features and $\vect{a}_{uv}$ is the feature of the edge $(u,v)$. The messages are aggregated at the target vertex $u$ using an operation of the form $\bigoplus_{v\in N(u)} \phi(\vect{x}_u, \vect{x}_v, \vect{h}_{uv})$. The functions $\bigoplus$, $\phi$, and $\psi$ change based on the high-level algorithm.
    The entire process of message generation and aggregation can be mapped to the flexible FusedMM operation developed in this paper.  
    }
    \label{fig:motivation}
    \vspace{-0.45cm}
\end{figure*}

{\em Message passing} is a powerful paradigm for designing various graph analytics and learning algorithms.
Given a graph $G(V,E)$ with $\vect{x}_u$ denoting node attributes for $u{\in}V$ and $\vect{a}_{uv}$ denoting edge attributes for $(u,v){\in}E$, a message passing system typically operates in two phases: (a) a message $\vect{h}_{uv}$ is generated on each edge $(u,v){\in}E$ using a function $\psi(\vect{x}_u, \vect{x}_v, \vect{a}_{uv})$ and (b) messages are aggregated at $u{\in}V$ using a function $\bigoplus_{v\in N(u)} \phi(\vect{x}_u, \vect{x}_v, \vect{h}_{uv})$.
Here, $N(u)$ denotes in-neighbors of $u$, $\psi$ and $\phi$ are application-defined functions and $\bigoplus$ is an application-defined aggregator. 
By changing the functions $\psi$, $\phi$, and $\bigoplus$, we can easily derive force-directed graph layout~\cite{jacomy2014forceatlas2, rahman2020batchlayout}, graph embedding~\cite{tsitsulin2018verse, force2vec}, graph convolutional network (GCN)~\cite{kipf2016semi}, and graph neural networks (GNNs) algorithms~\cite{xu2018powerful} as shown in Fig.~\ref{fig:motivation}.
This flexibility and intrepretability of message passing  made it a widely-used paradigm for designing high-level graph learning algorithms.

Even though a message passing API makes high-level graph algorithm easy to describe, high-performance linear algebra kernels are often used under the hood for performance.
Conceptually, the message generated on edges can be mapped to a sampled dense-dense matrix multiplication (SDDMM)~\cite{zhao2014high, nisa2018sampled,hong2019adaptive} and the message aggregation is performed by a sparse-dense matrix multiplication (SpMM).
For example, GNN frameworks such as PyTorch geometric (PyG)~\cite{fey2019pyg} and Deep Graph  Library (DGL)~\cite{wang2019dgl} rely on SDDMM and SpMM to implement their high-level message passing API.
When standard addition and multiplication operations can capture GNN's internal operations, PyG and DGL rely on vendor-provided sparse
libraries (e.g., MKL for Intel CPU, cuSPARSE for GPU) that offer highly
optimized implementations.
However, to capture complex and diverse operations such as those shown in Fig.~\ref{fig:motivation}, PyG and DGL also support user-defined operations using general SDDMM and SpMM kernels.

In almost all applications (except in attention-based GNNs~\cite{velivckovic2017graph}), messages generated on edges are immediately aggregated on vertices.
Computationally, it means that an SDDMM is almost always followed by an SpMM operation.
Existing libraries such as DGL provide separate SDDMM and SpMM kernels, forcing applications to generate intermediate outputs from SDDMM. 
This can incur significant computational and memory bottlenecks, especially when each edge generates high-dimensional messages. 
To address this problem, we develop a unified kernel called \emph{FusedMM} that captures the overall computation offered by SDDMM and SpMM. 
Conceptually, the fused kernel generates and aggregates messages collectively without explicitly storing messages. 
Thus, the updated feature $\vect{z}_u$ of vertex $u$ is generated in one shot as follows:
\vspace{-14pt}
\begin{equation}
   \vect{z}_u = \bigoplus_{v\in N(u)} \phi(\vect{x}_u, \vect{x}_v, \psi(\vect{x}_u, \vect{x}_v, \vect{a}_{uv})).
   \vspace{-4pt}
   \label{eq:fused_nodelevel}
\end{equation}
We develop efficient parallel algorithms for the FusedMM kernel that computes  Eq.~\ref{eq:fused_nodelevel} for every vertex. 
We show that this single kernel with the support of user-defined operations can capture almost all computational patterns arisen in graph layout, graph embedding and GNN algorithms. 
We make FusedMM general purpose by dividing it into five steps where each step performs vectorized operations with user-supplied functions.
We integrated FusedMM with DGL and show that even a naive implementation of FusedMM is always faster than DGL's SDDMM and SpMM operations on Intel, AMD and ARM processors.

For sparse graphs, FusedMM (as well as SDDMM and SpMM) is expected to be bound by the memory bandwidth.
Keeping this in mind, we developed a multithreaded algorithm that
minimizes data movements from the main memory and utilizes cache and registers as much as possible. 
To achieve best performance, we developed a library with code generator 
and tuned the factor of the register blocking after applying different 
strategies. Based on these tuned parameters, we automatically generated SIMD vectorized kernels with best register blocking supported on different SIMD architecture (e.g., AVX512/AVX in X86 and ASIMD/NEON in ARM). 
The optimized FusedMM is an order of magnitude faster than its equivalent kernels  in  DGL on Intel, AMD and ARM processors.
FusedMM  speeds  up  end-to-end  graph  embedding algorithms  by  up  to  $28\times$.
The main contributions of the paper are summarized below.
\begin{enumerate}[itemsep=0pt,topsep=4pt, leftmargin=4mm]
    \item We introduce FusedMM, a general-purpose kernel for various graph embedding and GNN operations. 
    \item FusedMM requires less memory and utilizes memory bandwidth efficiently by fusing SDDMM and SpMM operations.
    \item FusedMM employs autotuned vectorized operations that run up to 
    $34\times$
    faster than equivalent kernels in DGL. FusedMM performs equally well on Intel, AMD, and ARM processors.   
    \item FusedMM expedites end-to-end training of a graph embedding algorithm by $28\times$ relative to DGL.
\end{enumerate}

\Wskip{about accessing the output matrix once: it is true when $DIM <= bestDIM$ for our current version. The value of $bestDIM$ depends on hardware (e.g., on AVX512, it can be as large as 256 for single precision, may be 64 on ARM). To make the register-blocked version at least as good as unblocked when DIM is very large, we used register blocking for the first $bestDIM$ dimension only and kept remaining dimension unblocked in current version. We had to access some (unblocked) elements of output matrix multiple times for this case. If we want to access the output matrix only once by blocking, we need to access (each row) sparse matrix multiple times and will have redundant computations. Due to this trade of, it may become slower than the unblocked version for very large DIM. Nonetheless, it may be better to do an experiment and show the result and come to the conclusion of the current version ultimately.} 


\vspace{-5pt}
\begin{figure}
    \centering
    \vspace{-5pt}
    \includegraphics[width=\linewidth]{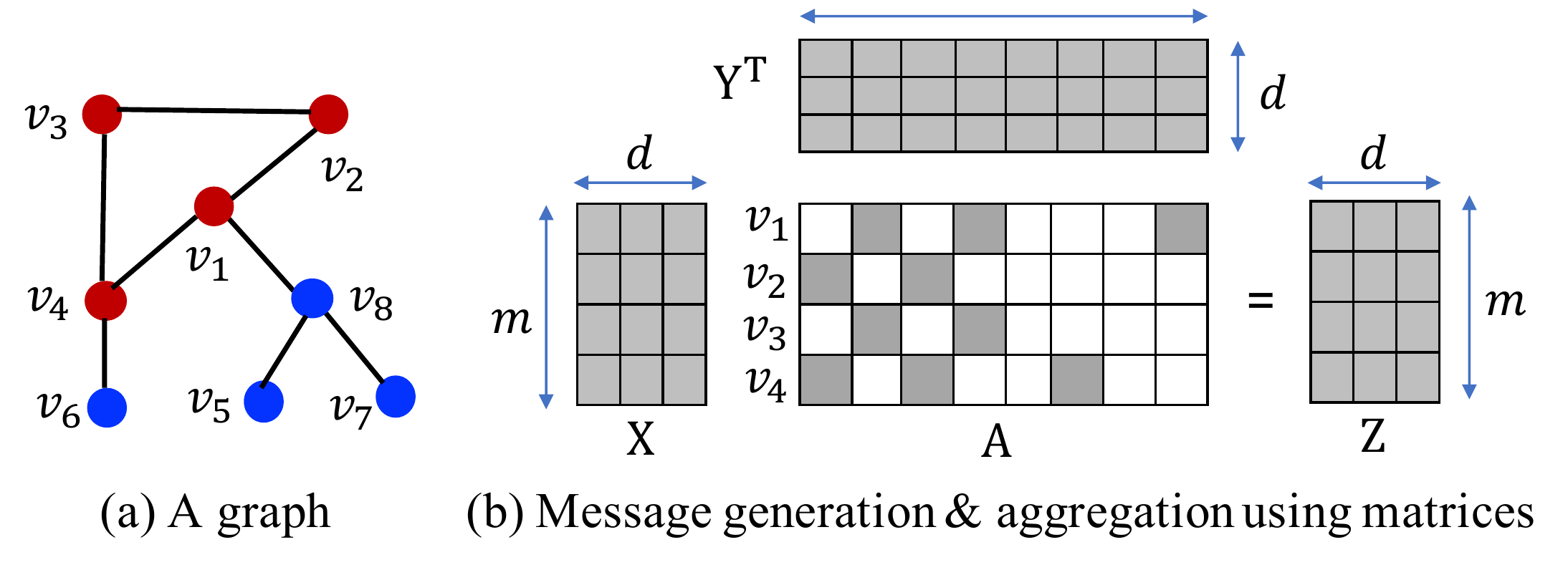}
    \vspace{-15pt}
    \caption{(a) An example graph. We consider message aggregations on a subgraph induced by $\{v_1, v_2, v_3, v_4\}$ (shown in red). (b) $\vect{A}$ denotes the adjacency matrix of the induced subgraph, $\vect{X}$ denotes features of $\{v_1, v_2, v_3, v_4\}$ and $\vect{Y}$ denotes features of all vertices. The updated features of $\{v_1, v_2, v_3, v_4\}$ are stored in $\vect{Z}$. 
    }
    \label{fig:settingLA}
    \vspace{-8pt}
\end{figure}

\vspace{-5pt}
\section{Linear-algebraic Kernels in Graph Learning}
\vspace{-2pt}
{\bf Notations.}
We use uppercase boldfaced letters to denote matrices.
$\vect{A}$ denotes the adjacency matrix,  $\vect{X}$ denotes 
features of the current subset of vertices, $\vect{Y}$ denotes  feature of all vertices, and $\vect{Z}$ denotes updated features of the current subset of vertices.
We use lowercase boldfaced letters to denote vectors.
The $u$th row of $\vect{A}$, $\vect{X}$, $\vect{Y}$, and $\vect{Z}$ are denoted by $\vect{a}_u$, $\vect{x}_u$, $\vect{y}_u$, and $\vect{z}_u$, respectively.
Hence, $\vect{x}_u{=}\vect{X}[u,:]$ represents the feature vector of the vertex $u$.
The feature of the edge $(u,v)$ is denoted by $\vect{a}_{uv}{=} \vect{A}[u,v]$.
Table~\ref{tab:notations} summarizes our notations. 

\begin{table}
\centering
\caption{List of notations used in the paper}
\vspace{-3pt}
\begin{tabular}{l l} 
\toprule
\textbf{Symbol}                 & \textbf{Description}                         \\ 
\toprule
$\vect{A}$ & A sparse matrix with dimension: $m \times n$        \\ 
$m$                             & The number of rows in $\vect{A}$                      \\ 

$n$                             & The number of columns in $\vect{A}$                  \\ 

$\nnz(\vect{A})$                           & The number of non-zero elements in $\vect{A}$         \\ 

$d$                             & The dimension of embedding                   \\ 
$\vect{X}$ & A dense input matrix with dimension: $m \times d$   \\ 

$\vect{Y}$ & A dense input matrix with dimension: $n \times d$   \\ 

$\vect{Z}$ & A dense output matrix with dimension: $m \times d$  \\

$\vect{A} \times \vect{B}$ & Matrix-matrix multiplication \\

$\vect{A} \odot \vect{B}$ & Element-wise multiplication \\
$\vect{a}_{uv} = \vect{A}[u,v]$ &  features of the edge $(u,v)$ \\
$\vect{x}_u = \vect{X}[u,:]$ &  $d$-dimensional feature vector of vertex $u$\\ 
$\vect{a}_u = \vect{A}[u,:]$ &  $u$th row of the adjacency matrix \\
&  storing edges adjacent to $u$\\ 

\bottomrule
\end{tabular}

\label{tab:notations}
\vspace{-0.55cm}
\end{table}

{\bf The problem setting.}
Let $G(V,E)$ denote a graph with a set of $n$ vertices $V$ and a set of edges $E$.
In most practical settings, an induced subgraph (e.g., a minibatch of vertices) is considered at a given point. 
For example, vertices $\{v_1, v_2, v_3, v_4\}$ and their adjacent edges form a minibatch in Fig.~\ref{fig:settingLA}(a).
We consider developing a linear-algebra kernel that can capture both message generation and aggregation for all vertices in a given subgraph (also covers the case for the entire graph).

Let $\vect{A} \in \mathbb{R}^{m\times n}$ denote the sparse adjacency matrix of the given subgraph of $m$ vertices where $\vect{a}_{uv} \neq 0$ if $(u,v){\in}E$, otherwise $\vect{a}_{uv} = 0$.
Here, the rectangular matrix $\vect{A}$
represents a slice of the original adjacency matrix and thereby captures a minibatch of vertices needed in GNN training.
Similarly, $\vect{A}$ can also represent a bipartite graph with different number of vertices in its two parts. 
$\vect{a}_{uv}$ can be either  Boolean (unweighted graphs) or a user-defined data type depending on edge features. 

Let $\vect{X} {\in} \mathbb{R}^{m\times d}$ be the dense matrix storing features of vertices in the current subgraph and $\vect{Y} {\in} \mathbb{R}^{n\times d}$ be a feature matrix for all vertices. 
Each row of $\vect{X}$ or $\vect{Y}$ stores a $d$-dimensional vertex feature vector. 
Even though $\vect{X}$ can be a submatrix of $\vect{Y}$ in most practical applications, we separate them for generality. For example, $\vect{X}$ and $\vect{Y}$ may store different features in heterogeneous and bipartite graphs. 
The final output of a message passing step is an updated feature matrix $\vect{Z} {\in} \mathbb{R}^{m\times d}$, where the $u$th row $\vect{z}_u$ stores the updated features of $u$.
We seek a fused matrix multiplication kernel FusedMM that outputs $\vect{Z}$ from $\vect{A}, \vect{X}$, and $\vect{Y}$ as follows $\vect{Z}{=}\text{FusedMM}(\vect{A},\vect{X},\vect{Y})$.
Fig.~\ref{fig:settingLA}(b) shows the linear-algebraic representation.

\begin{figure*}[!t]
    \centering
    \includegraphics[width=.98\linewidth]{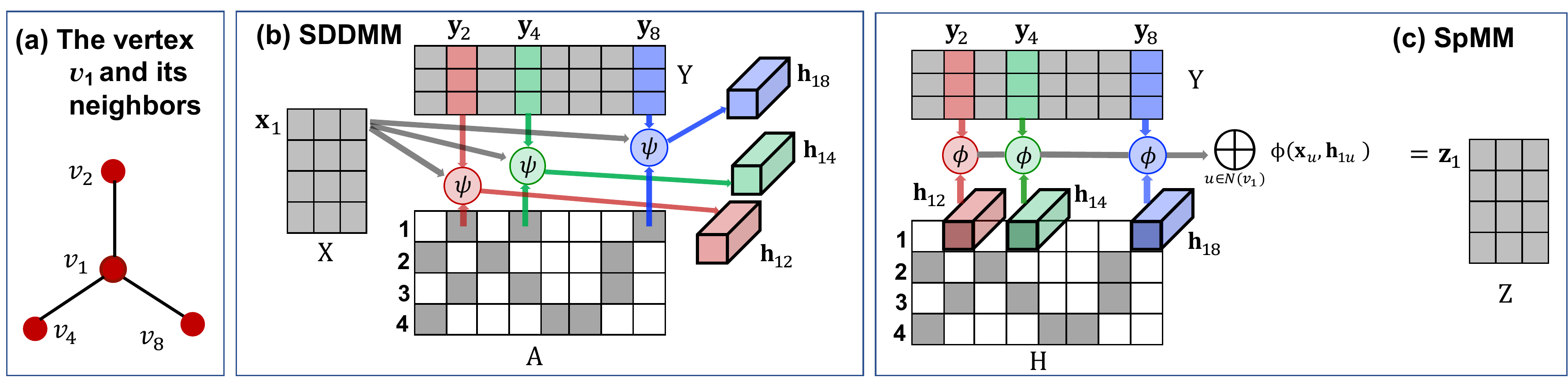}
    \vspace{-5pt}
    \caption{(a) A vertex $v_1$ and its neighbors. We would like to generate messages on all edges adjacent to $v_1$ and then aggregate the messages to get new feature vector for $v_1$. (b) $\vect{x}_1$ denotes the feature vector of $v_1$. $\vect{y}_2$, $\vect{y}_4$, and $\vect{y}_8$ denote feature vectors of $v_1$'s neighbors  $v_2$, $v_4$, and $v_8$, respectively.  An SDDMM is used to generate messages $\vect{h}_{12}, \vect{h}_{14}$, and  $\vect{h}_{18}$ for the edges adjacent to $v_1$. Here, $\vect{h}_{12}, \vect{h}_{14}$, and  $\vect{h}_{18}$ can be vectors. (c) The messages are aggregated using an SpMM operation that generates the updated vector $\vect{z}_1$ for $v_1$. When SDDMM and SpMM separated as shown here, the messages are explicitly stored in $\vect{H}$. This paper combines SDDMM and SpMM into a FusedMM operation.}
    \label{fig:messagepassing}
\end{figure*}

Even though a FusedMM operation is what most applications need, PyG and DGL map edge-wise computations to an SDDMM kernel and vertex-wise computations to an SpMM kernel.
We briefly discuss how these separate kernels are used.

{\bf Edge-wise message kernel: SDDMM.}
As shown in Fig.~\ref{fig:motivation}, messages are generated on an edge $(u,v)$ based on features of vertices  $u$ and $v$.
When we consider messages generated on all edges corresponding to $\vect{A}$, the underlying operation is an SDDMM operation that operates on two dense input matrices guided by a sparse matrix, and produces a sparse matrix $\vect{H}$: 
\vspace{-3pt}
\begin{equation}
\label{eqn:sddmmout}
   \text{[SDDMM]\ \ \ \ \ } \vect{H} = (\vect{X\times Y}^{T})\bigodot \vect{A}.
   \vspace{-3pt}
\end{equation}
Here, $\vect{H}$ is an $m\times n$ sparse matrix or tensor with exactly the same sparsity pattern of $\vect{A}$, and $\bigodot$ represents element-wise multiplication.
Since $\vect{A}$ is a sparse matrix, any practical implementation of SDDMM would only compute entries where $\vect{A}$ has nonzeros as shown in Fig.~\ref{fig:messagepassing}(b). 
The actual computation needed to generate $\vect{H}[u,v]{=}\vect{h}_{uv}$ is application dependent and is represented by the message generation function $\psi$. 
Hence, in a general-purpose SDDMM (gSDDMM), we 
have $\vect{h}_{uv} {=} \psi(\vect{x}_u, \vect{y}_v, \vect{a}_{uv}$), 
where $\vect{x}_u$ and $\vect{y}_v$ denote the feature vectors of $u$ and $v$.
Notice in  Fig.~\ref{fig:messagepassing}(b) that $\psi$ can generate a vector as an output making $\vect{H}$ a sparse tensor in some applications. 
At the end of the gSDDMM operation, $\vect{h}_{uv}$ stores the message generated on the edge $(u,v)$. 


{\bf Vertex-wise message aggregation kernel: SpMM.}
After generating messages, most applications aggregate them on target vertices. This operation can be captured by an SpMM operation: \vspace{-9pt}
\begin{equation}
  \text{[SpMM]\ \ \ \ \ } \vect{Z} = \vect{H}\times \vect{Y},
  \vspace{-3pt}
\end{equation}
where each row of $\vect{Z}$ stores the updated feature vectors of vertices.
As with the gSDDMM operation, a generalized SpMM (gSpMM) can take user-defined multiplication operation $\phi$ and aggregation function as shown in Fig.~\ref{fig:messagepassing}(c).
Thus, the $u$th row of the output can be formed as follows:
$\vect{z}_u = \bigoplus_{\vect{h}_{uv}\neq 0} \phi(\vect{y}_{v}, \vect{h}_{uv})$, 
where $\vect{z}_u=\vect{Z}[u,:]$ denotes the updated feature vector of the target vertex $u$.

{\bf The need for a fused matrix-multiplication kernel.} 
The first and most obvious reason to develop a unified kernel is to reduce the memory requirement. 
For example, the intermediate matrix $\vect{H}$ storing edge-wise messages can take $O(d*\nnz(\vect{A}))$ space when the application generates $d$-dimensional messages on edges. 
This extra memory required to store $\vect{H}$ can make it downright impossible to solve certain problems if we use separate SDDMM and SpMM kernels. 
Separate kernels also require us to read/write $\vect{H}$ and $\vect{Y}$ more than once, which 
negatively impacts the performance of these memory-bound kernels.
Finally, a unified kernel enables a joint optimization of message generation and aggregation, offering more flexibility to applications and more opportunity to tune performance on different processors and accelerators.   
Motivated by these potential benefits in multiple frontiers, we develop FusedMM that offers both flexibility and high performance. 



\section{A Flexible FusedMM Kernel}
{\bf Design objectives.}
The first objective of the FusedMM kernel is to capture the entire message generation and aggregation process for a given subgraph as formulated in Fig.~\ref{fig:settingLA}(b).
The main challenge in designing such a unified kernel is to make it flexible for diverse applications that use various message generation functions $\psi$, multiplication operations $\phi$, and message aggregators $\oplus$. 
We address this challenge by splitting the whole computation into a sequence of five well-defined steps where each step accepts user-defined functions. 
The second objective of FusedMM is to efficiently utilize processor resources (registers, cache, memory bandwidth, etc.) to maximize performance.   
We achieve this objective by fully utilizing SIMD units available in modern processors while maximizing the utilization of the memory bandwidth. 

\vspace{-5pt}
\subsection{The anatomy of FusedMM}
\vspace{-3pt}
\label{sec:abstactop}
We begin with a graph embedding example shown in Fig.~\ref{fig:motivation}(b).
The message $\vect{h}_{uv}$ generated on edge $(u,v)$ is defined by $\sigma(\vect{x}_u^T\vect{y}_v)$, where $\sigma$ is the Sigmoid function. As mentioned before, we separate features of source and destination vertices for generality.
We split the computation of $\vect{h}_{uv}$ into three steps. 

    {\bf [Step 1: VOP]}  We perform elementwise ``multiplication" of  $\vect{y}$ and $\vect{x}$ and return another vector $\vect{w}$ of the same length of $\vect{y}$ and $\vect{x}$. For our graph embedding example, this step computes the first part of the dot product. As this step operates on two vectors, we call it VOP. More specifically, $\text{VOP}(\vect{x},\vect{y}) = \vect{x} \odot \vect{y} = \vect{z}$, where $\odot$ is an element-wise vector operation. Any user-defined function following this syntax is also allowed.    
    
    {\bf [Step 2: ROP]} The second step potentially reduces a vector to a scalar. For our graph embedding example, this step computes the summation part of the dot product computation. As this step reduces elements of a vector, we call it ROP. More specifically, $\text{ROP}(\vect{z})= \oplus_i \vect{z}_i = s$, where $s$ is a scalar and $\oplus$ is a reduction operation. VOP and ROP together can compute the dot product between two vectors.   
    
    {\bf [Step 3: SOP]} The third step scales a vector/scalar using a linear or nonlinear function. For our graph embedding example, this step applies the sigmoid function to $\vect{x}_u^T\vect{y}_v$, hence it is called a scaling operation or SOP. More specifically, $\text{SOP}(\vect{z})= \sigma(\vect{z}_i); \forall i$, where $\sigma$ is a linear or nonlinear function. $\vect{z}$ can also be scalar such as in computing $\sigma(\vect{x}_u^T\vect{y}_v)$.

These three steps deliver a flexible message-generation operation $\phi$ and form the SDDMM phase of FusedMM. 
Next, we consider the message aggregation phase in graph embedding shown in the Fig.~\ref{fig:motivation}(b).
The message aggregation is defined by:
$\vect{z}_u {=}\sum_{v\in N(u)} \vect{h}_{uv}{\odot} \vect{y}_v$, where $\vect{h}_{uv}$ is the output of SDDMM on the edge $(u,v)$.
$\vect{h}_{uv}$ is either a scalar or a vector. 
We split the message aggregation into two steps.

    {\bf [Step 4: MOP]}  We perform elementwise ``multiplication" of  $\vect{h}$ and $\vect{y}$ and return another vector  $\vect{w}$ of the same length of $\vect{y}$. If either input is a scalar (e.g., the message generated on an edge is scalar),  MOP simply scales the other vector with the scalar.
    For graph embedding where $\vect{h}_{uv}$ is a scalar, this operation scales $\vect{y}_v$ by $\vect{h}_{uv}$. As this step multiplies a vector by a vector or scalar, we call it MOP. Specifically, $\text{MOP}(\vect{h},\vect{y}) = \vect{w} = \vect{h} \odot \vect{y}$, where $\odot$ is an element-wise ``multiplication" operation. 
    
    {\bf [Step 5: AOP]} The last step ``accumulates" a vector with another vector.  For graph embedding, this step accumulates messages received from adjacent vertices; hence it is called an accumulation operation or AOP. Specifically, $\text{AOP}(\vect{z,y}) = \vect{z} = \vect{z}\oplus\vect{y}$, where $\oplus$ is a use-defined ``addition" function. 

All of these five steps together form the core computations of FusedMM.
We define these operation in an abstract sense so that users can use standard or custom-built operations to substitute them.
Our library accepts function pointers for each of these five operations to facilitate user-defined functions. 
An application can skip some of these steps by passing a NOOP.

The aforementioned breakdown serves two key objectives of FusedMM.
First, these steps are flexible enough to capture almost all operations in various graph layout, graph embedding and GNN algorithms (see Table~\ref{tab:fusedMM-operation}).
By changing the operations, users can design their own high-level applications.
Second, all five operations are similar to level-1 BLAS operations that can be vectorized using SIMD units.

\begin{table}[!t]
\centering
\caption{Standard operations that could be used in five steps of FusedMM. $\vect{x}, \vect{y}, \vect{z}$ are vectors,  $\alpha$ is a scalar, $\sigma$ is a unary function. $\oplus$ and $\odot$ denote ``addition" and ``multiplication" operations, respectively. In the most general case,  $\oplus$ and $\odot$ are defined as part of a semiring provided by the application. $^1$\footnotesize{Our ASUM is different from the absolute sum of L1 BLAS.}}
\vspace{-3pt}
\begin{tabular}{ l p{1.0cm} p{0.5cm} p{0.5cm} p{1.6cm} m{1.3cm}}
\toprule
\textbf{Type} & \textbf{Op}                    & \textbf{In 1} & \textbf{In 2} & \textbf{Return} & \textbf{Where used}  \\
\toprule
& ADD & $\vect{x}$ & $\vect{y}$ & $\vect{x} \oplus \vect{y}$ & VOP\\
Binary & MUL & $\vect{x}$ & $\vect{y}$ & $\vect{x} \odot \vect{y}$ & VOP, MOP\\
& SEL2ND & $\vect{x}$ & $\vect{y}$ & $\vect{y}$ & VOP, MOP\\
\midrule
Unary & SIGMOID & $\vect{x}$ & $\sigma$ & $\sigma(\vect{x})$ & SOP, MOP\\
 & SCAL & $\vect{x}$ & $\alpha$ & $\alpha\vect{x}$ & SOP, MOP\\
\midrule
Reduction & RSUM & $\vect{x}$ & - & $\oplus_i \vect{x}_i$ & ROP\\
& RMUL & $\vect{x}$ & - & $\odot_i \vect{x}_i$ & ROP\\
\midrule
Accumulate & ASUM$^1$ & $\vect{z}$ & $\vect{x}$ & $\vect{z} {\gets} \vect{z} \oplus \vect{x}$ & AOP\\
 & AMAX & $\vect{z}$ & $\vect{x}$ & $\vect{z} {\gets} \max(\vect{z}, \vect{x})$ & AOP\\
\bottomrule
\end{tabular}
\label{tab:standard-ops}
\vspace{-0.2cm}
\end{table}

\subsection{Standard operations used in FusedMM with applications}
\vspace{-2pt}
While designing FusedMM with VOP, ROP, SOP, MOP, and AOP makes a very general kernel, users have to provide definitions of these operations to design a new application.
To this end, we develop optimized implementations for a few common operations that users can directly plug into their applications.
Table~\ref{tab:standard-ops} shows some standard operations that we implemented in our software. We also show where these operations could be used inside FusedMM.
For example, binary operations ADD and MUL denote element-wise addition and multiplication operations that could be used to substitute VOP and MOP. 
The SEL2ND operations simply copies the second operand to the output.
Unary operations such as SIGMOID and SCAL are used to scale a vector using non-linear and linear functions. 
Hence, these operations can be used in place of SOP and MOP in FusedMM.
Finally, reduction operations RSUM and RMUL are used in ROP, and accumulation operations ASUM and AMAX are used in AOP.
Note that Table~\ref{tab:standard-ops} only shows a few examples that can be used in FusedMM.
Any user-defined functions are allowed as long as they satisfy the I/O requirements of VOP, ROP, SOP, MOP, and AOP.


\begin{table*}[!t]
\centering
\caption{Using FusedMM to develop four applications. NOOP means no operation needed at that step. $^1$ \small{Needs a user-provided MLP function}. $^2$ \small{A function that computes vector norms.}}
\vspace{-3pt}
\begin{tabular}{ l c c c c c c }
\toprule
\textbf{Application}                    & \textbf{Ref} & \textbf{VOP} & \textbf{ROP} & \textbf{SOP} & \textbf{MOP} & \textbf{AOP} \\ 
\toprule
Graph Layout (FR model~\cite{jacomy2014forceatlas2}) & Fig.~\ref{fig:motivation}(a)            & ADD          & NORM$^2$   & SCAL   & MUL          & ASUM          \\ 
Node embedding (Force2Vec~\cite{force2vec} and VERSE~\cite{tsitsulin2018verse} with sigmoid)  & Fig.~\ref{fig:motivation}(b)            & MUL    & RSUM          & SIGMOID   & MUL          & ASUM          \\  
                                 

Graph Convolution Network~\cite{kipf2016semi}      & Fig.~\ref{fig:motivation}(c)            & SEL2ND    & NOOP         & NOOP         & MUL         & ASUM          \\ 

Graph Neural Network with MLP     & Fig.~\ref{fig:motivation}(d)            & MLP$^1$    & NOOP         & SIGMOID         & MUL         & AMAX          \\

                                   \bottomrule
\end{tabular}
\label{tab:fusedMM-operation}
\vspace{-10pt}
\end{table*}

Table~\ref{tab:fusedMM-operation} shows how we can implement four applications described in Fig.~\ref{fig:motivation}. 
We already discussed the graph embedding application when we defined VOP, ROP, SOP, MOP, and AOP in Section~\ref{sec:abstactop}.
For GCN as shown in Fig.~\ref{fig:motivation}(c), the first VOP operation simply selects neighbor's feature using SEL2ND.
Since a vanilla GCN does not perform a reduction on edges, we use NOOP for ROP and SOP in the 2nd row in Table~\ref{tab:fusedMM-operation}.
The message aggregation in GCN multiplies messages by edge features using MUL for MOP and finally messages are pooled using ASUM. Different variants of GCN use different pooling options~\cite{xu2018powerful} such as maximum, minimum, mean, etc.
All of these options can be captured by MOP and AOP in FusedMM.
The fourth row in Table~\ref{tab:fusedMM-operation} shows a simple GNN layer that uses MLP to generate messages. 
This is an example where a user-defined VOP is needed.

\setlength{\textfloatsep}{0pt}
\begin{algorithm}[!t]
\caption{The FusedMM algorithm}\label{algo:fusedmmstart}
\textbf{Input:} $\vect{A}$: the adjacency matrix, $\vect{X}$: the dense embedding matrices of dimension $m {\times} d$,
$\vect{Y}$: the dense embedding matrices of dimension $n {\times} d$
\textbf{Output:} $\vect{Z}$: an $m \times d$ matrix 
\begin{algorithmic}[1]
\Procedure{\text{FusedMM}}{$\vect{A}, \vect{X}, \vect{Y}$}
    \State $\{\vect{A}_1,..., \vect{A}_t\} \gets $ \textproc{Part1D}($\vect{A}$)  \Comment{$\nnz(\vect{A}_i)
    {\approx} \frac{1}{t}\nnz(\vect{A})$}
    \State $\{\vect{X}_1,..., \vect{X}_t\} \gets $ \textproc{Part1D}($\vect{X}$)  \Comment{$\text{nrow}(\vect{X}_i)
    {=} \text{nrow}(\vect{A}_i)$}

    \For{$i \in {1..t}$ {\bf in parallel}} \Comment{Thread parallel}
        \For{ each row $u$ of $\vect{A}_i$ } \Comment{Iterate over rows}
            \State $\vect{x}_u \gets \vect{X}_i[u,:] $\ \  \ \ $\vect{a}_u \gets \vect{A}_i[u,:] $
            \State $\vect{z}_u \gets$\textproc{UpdateU}($\vect{a}_u, \vect{x}_u, \vect{Y}$) 
            
            
         \EndFor
    \EndFor
\State \Return $\vect{Z}$
\EndProcedure

\Procedure{\text{UpdateU}}{$\vect{a}_u, \vect{x}_u, \vect{Y}$} \Comment{Message generation and aggregation for the vertex $u$}
            \State $\vect{z}_u \gets 0$            
            \For{ each $v$ with  $\vect{a}_{uv}\neq 0$ }
            \State $\vect{y}_v \gets \vect{Y}[v,:] $ 
            \State $\vect{z} \gets $VOP($\vect{x}_u$, $\vect{y}_v$) 
            \State $s\gets $ ROP($\vect{z}$)             
            \State $\vect{h} \gets $ SOP($s$ or $\vect{z}$)   \Comment{directly use  $\vect{z}$ if ROP is a NOOP, otherwise use $s$}
            \State $\vect{w} \gets $ MOP($\vect{h}, \vect{y}_v$)  
            \State $\vect{z}_u \gets$  AOP($\vect{z}_u, \vect{w}$)
            
        \EndFor

\State \Return $\vect{z}_u$
\EndProcedure
\end{algorithmic}
\label{algo:fusedmm}
\vspace{-0.1cm}
\end{algorithm}
\setlength{\textfloatsep}{10pt plus 1.0pt minus 2.0pt}

\vspace{-2pt}
\subsection{The parallel FusedMM algorithm}
\vspace{-3pt}
Using all building blocks of FusedMM discussed thus far,  
Algorithm~\ref{algo:fusedmm} describes a multithreaded FusedMM algorithm that takes $\vect{A}, \vect{X}$, and  $\vect{Y}$ as inputs and returns updated vertex features $\vect{Z}$ as the output. 
FusedMM does not perform minibatching, which is done at the application layer.

{\bf Partitioning matrices for thread-level parallelization.}
Parallel FusedMM in Algorithm~\ref{algo:fusedmm} starts with load-balanced partitioning of matrices. 
We partition input matrices with two key objectives in mind: (a) computations in different parts should be independent of each other so that threads can process different partitions in parallel without synchronization and (b) the computational cost for each partition should be approximately equal.
If we aim to implement SDDMM and SpMM separately, we could use either 1D (i.e., vertex partitioning) or 2D (edge partitioning) partitioning of the adjacency matrix $\vect{A}$.
However, when we fuse SDDMM and SpMM in FusedMM,  2D partitioning of $\vect{A}$ may be very inefficient or even outright impossible.
In our graph embedding example, the message on edge $(u,v)$ is computed using a dot product of node feature vectors and then, a sigmoid function is applied on the output of the dot product.
In this case, it is not possible to generate messages from partial vertex features without changing the mathematical interpretation. 
Even when partial messages make sense, 2D  partitioning may be very inefficient because the last step of FusedMM aggregates messages from in neighbours $N(u)$ of $u$, but all vertices in $N(u)$ may not be in the current partition.
Consequently, 2D partitioning will require storing partially computed results, which could be detrimental for performance. Thus, we opt to use 1D partitioning of $\vect{A}$. 

\begin{figure}[!t]
    \centering
    \vspace{-2pt}
    \includegraphics[width=.87\linewidth]{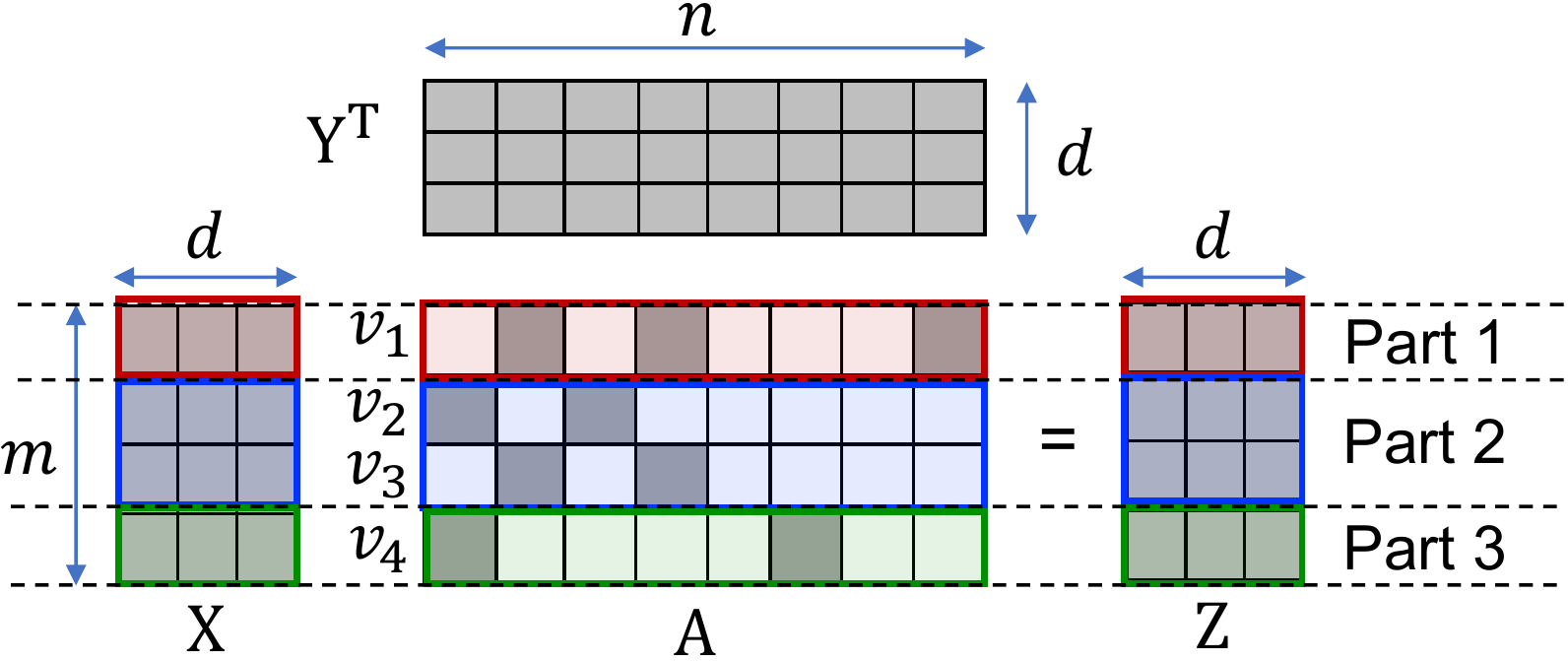}
    \vspace{-3pt}
    \caption{1D partitioning of $\vect{X}, \vect{A}$ and $\vect{Z}$ for thread-level parallelization. Three partitions are shown in three colors. For example,  Algorithm~\ref{algo:fusedmm} uses $\vect{A}_2$ to denote the middle partition (shown in blue) of $\vect{A}$. 
    }
    \label{fig:partition}
    \vspace{-3pt}
\end{figure}

The \textproc{Part1D} function at line 2 of Algorithm~\ref{algo:fusedmm} partitions $\vect{A}$ into $t$ parts $\vect{A}_1,..., \vect{A}_t$, where each part has the same number of columns but different numbers of rows. 
We use a simple load-balancing scheme where $\nnz$ of all parts $\vect{A}_i$ are approximately equal: $\nnz(\vect{A}_i) {\approx} 1/t*\nnz(\vect{A})$.
This partitioning is done by scanning the row pointer array of $\vect{A}$ stored in the Compressed Sparse Row format.
The complexity of \textproc{Part1D} is $O(m)$, where $m$ is the number of rows in $\vect{A}$.
$\vect{X}$ and $\vect{Z}$ are partitioned following the partitions of $\vect{A}$ such that $\text{nrow}(\vect{X}_i){=} \text{nrow}(\vect{A}_i){=}\text{nrows}(\vect{Z}_i)$.
The other input matrix $\vect{Y}$ is not partitioned.
Fig.~\ref{fig:partition} shows an example of 1D partitioning. 

{\bf Fused operation within a thread.}
After the partitioning is performed, the $i$th group of partition  $\{\vect{A}_i, \vect{X}_i, \vect{Y}\}$ are processed by a thread to generate the corresponding output $\vect{Z}_i$. 
Here, threads may perform concurrent reads on $\vect{Y}$, but they do not perform any concurrent writes. 
Hence, threads can proceed in parallel (line 5-7 of Alg.~\ref{algo:fusedmm}) without any synchronization.  

Within its private data partitions, a thread   processes one vertex (that is one row of $\vect{A}_i$) at a time (line 5 of Alg.~\ref{algo:fusedmm}).
Assume that $u$ is the current vertex under consideration.
Then, the \textproc{UpdateU} function generates messages for all edges adjacent to $u$ and aggregates the messages to return the updated feature vector $\vect{z}_u$.
Thus, the \textproc{UpdateU} function needs the entire $\vect{Y}$ and the $u$th rows of $\vect{X}$ and $\vect{A}$ to generate $\vect{z}_u$.
Inside the \textproc{UpdateU} function (line 9-18 of Alg.~\ref{algo:fusedmm}), we call five building blocks VOP, ROP, SOP, MOP, and AOP to obtain our desired vector $\vect{z}_u$.
In our library, \textproc{UpdateU} can take pre-defined or user-defined functions. 

\section{Optimizations and Code Generation}

\begin{figure}
    \centering
    \includegraphics[width=0.9\linewidth]{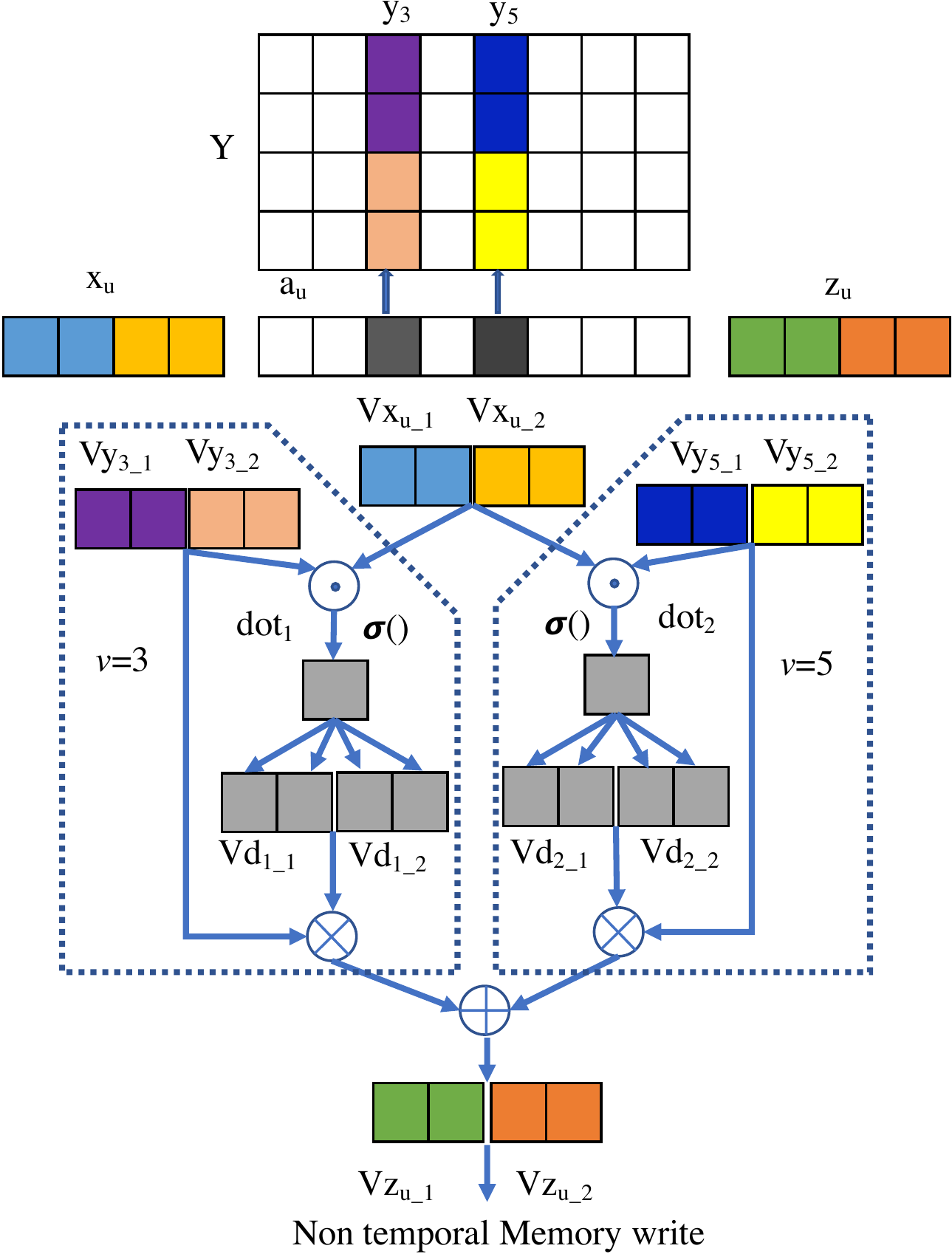}
    \caption{SIMD vectorized implementation of \textproc{UpdateU} function for specialized sigmoid based graph embedding. All the labels, starting with `V', indicate the SIMD  registers and they are color-coded with the portion of the vector they access or update.}
    \label{fig:simd-fusedMM}
    \vspace{-3pt}
\end{figure}

While the general FusedMM algorithm (Alg.~\ref{algo:fusedmm}) provides ample flexibility to applications, it can perform sub-optimally because it stores outputs after each of its five steps. 
If we recognize a pattern from predefined VOP, ROP, SOP, MOP, and AOP operations, we can optimize the whole kernel by feeding the output of one operation directly to the next operation without storing the results. For example, the second row in  Table~\ref{tab:fusedMM-operation} uses the (MUL, RSUM, SIGMOID, MUL, ASUM) sequence of operations, which is a common pattern used by various node embedding algorithms such as VERSE~\cite{tsitsulin2018verse} and Force2Vec~\cite{force2vec}.
Hence, we develop optimized FusedMM kernels for patterns denoted by the first three rows in Table~\ref{tab:fusedMM-operation}.
This allows us to optimize them using architecture-dependent intrinsic codes to be discussed in the next section. 

\vspace{-2pt}
\subsection{SIMD Vectorization and Register Blocking}
\vspace{-3pt}
While each thread computes the \textproc{UpdateU} operation in Algorithm~\ref{algo:fusedmm} for vertices of its own partition in a core, these computations can be further optimized by Single Instruction Multiple Data 
(SIMD) parallelism. By employing register blocking of $\vect{x}_u$, $\vect{y}_v$ and $\vect{z}_u$ in SIMD registers, we can effectively reduce the number of their accesses. Note that the same vectors $\vect{x}_u$ and $\vect{z}_u$ are accessed in line~13 and line~17 for all iterations of the loop in \textproc{UpdateU} in Algorithm~\ref{algo:fusedmm}. Therefore, we load $\vect{x}_u$ in SIMD registers and initialize the registers for $\vect{z}_u$ with zero before entering into the loop and use those registers inside for the intermediate computations throughout all iterations of the loop. 
We can avoid loading of $\vect{z}_u$ and use non-temporal writes from registers to the memory directly without polluting the cache in this way. Therefore, the register blocking of $\vect{x}_u$ and $\vect{z}_u$ will reduce the number of accesses (either in cache or memory) for those by the number of their neighbors (average degree of the graph). We can also register block $\vect{y}_v$ inside the loop at line~12. However, it will reduce the number of access for $\vect{y}_v$ by only a factor of two since we are accessing this vector in VOP (line~13) and MOP (line~16) operations, and each iteration of the loop will access a different $\vect{y}_v$ vector. 

Fig.~\ref{fig:simd-fusedMM}
shows an example of SIMD vectorized \textproc{UpdateU} operation of a specialized sigmoid-based graph embedding algorithm where the dimension $d$ is four and the width of the SIMD register is two. Note that the pattern of this algorithm is known to our library and it already has specialized implementation for this type of operation.   
At the beginning of the \textproc{UpdateU} function, we load $\vect{x}_u$  into two SIMD registers $\vect{Vx}_{u\_1}$ and $\vect{Vx}_{u\_2}$. We initialize the SIMD registers $\vect{Vz}_{u\_1}$ and $\vect{Vz}_{u\_2}$ with zero for the $\vect{z}_u$. We load each $\vect{y}_v$ into two registers in each iteration of the loop and dot-product them with $\vect{Vx}_{u\_1}$ and $\vect{Vx}_{u\_2}$ (VOP + ROP). The reduced scalar value from the dot product is scaled with sigmoid function (SOP) and broadcasted to SIMD registers. 
The registers $\vect{Vz}_{u\_1}$ and $\vect{Vz}_{u\_2}$ which hold the intermediate results for $\vect{z}_u$ are fused multiply-accumulated (FMAC) with those broadcasted values and the registers which hold the values of $\vect{y}_v$ (MOP + AOP). 
After accumulating all the values for all the neighbors into the $\vect{Vz}_{u\_1}$ and $\vect{Vz}_{u\_2}$, we store the values of these registers to memory only once after exiting from the loop and hence reduce the memory accesses for $\vect{z}_u$.

\begin{figure}[t]
    \centering
    \includegraphics[width=1.0\linewidth]{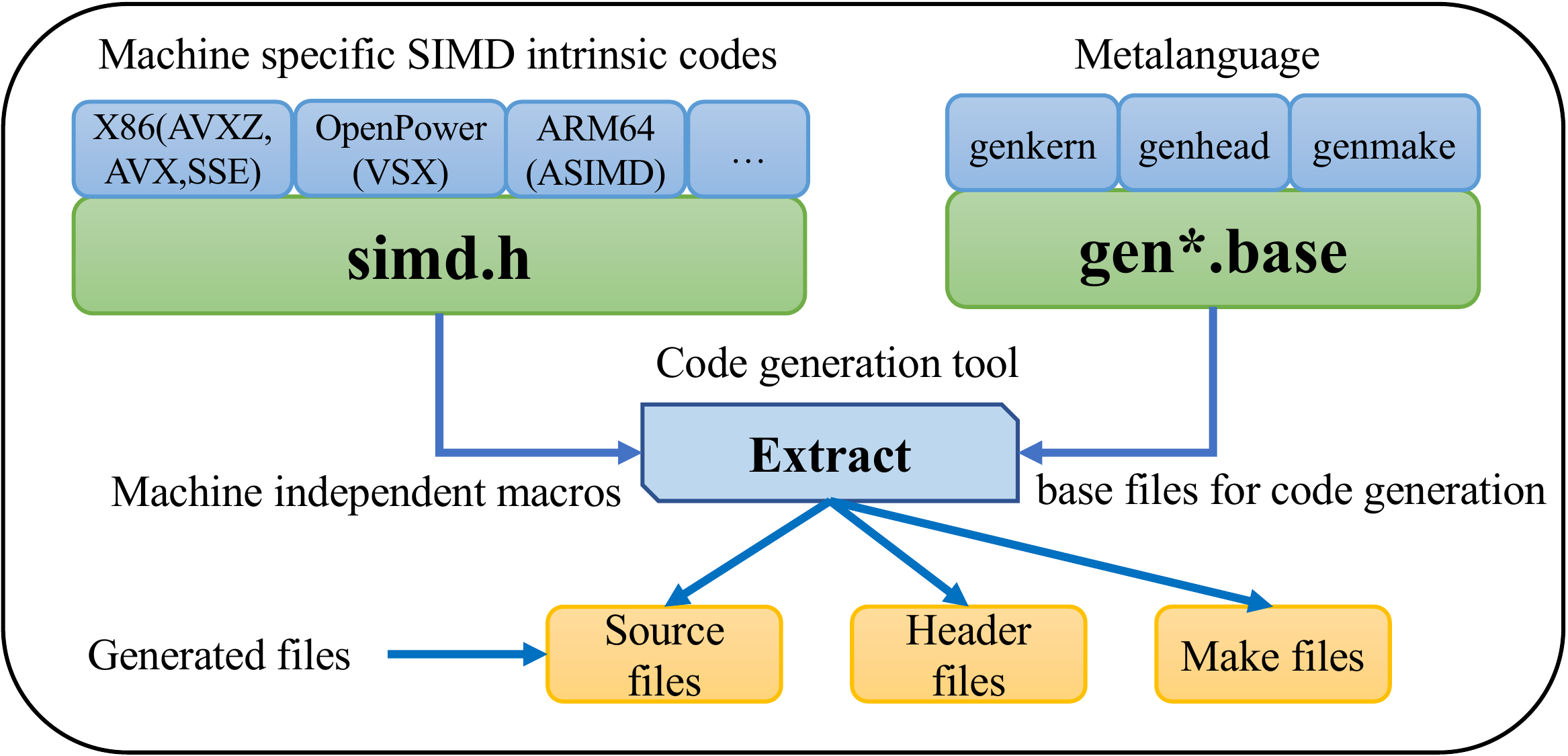}
    \caption{Block diagram for the SIMD code generation. Three different basefiles are used to generate source, header, and makefiles using {\em extract} tool. The header file {\em simd.h} is used to provide macros for SIMD operations and to hide architecture-specific intrinsic codes.}
    \label{fig:codegen}
    \vspace{-3pt}
\end{figure}

\vspace{-2pt}
\subsection{Code Generation}
\vspace{-3pt}
We use a code generation tool, \emph{extract}, from~\cite{atlas_sc98} to generate SIMD intrinsic codes on different hardware architectures. 
Fig.~\ref{fig:codegen} shows a block diagram of the structure of the code generator. We use metalanguage of \emph{extract} to write basefiles which are used to generate source codes, header files and makefiles in our library. The header file \text{simd.h} in Fig.~\ref{fig:codegen} is used to hide the hardware specific SIMD intrinsic codes and provides a common interface for all supported hardware using C preprocessor macros. Note that this design makes it very easy to add support for new SIMD instruction set architectures (ISA). 
We will only need to add the implementation of the macros using the intrinsic of new hardware supported by compilers. We parameterize the code generation process and generate kernels for predefined patterns of operations based on a range of dimension values, register blocking strategies, and data types. 
In our implementation, we can choose vectors, prioritize the output vectors over read-only vectors for register blocking, and even limit register blocking up to a threshold when the dimension is large.

\vspace{-2pt}
\subsection{Complexity and expected performance}
\vspace{-3pt}
As shown in Alg.~\ref{algo:fusedmm}, FusedMM uses five operations  in the \textproc{UpdateU} function.
Assuming each vector is of length $d$, the complexity of any of the five operations is $O(d)$.
As FusedMM calls these operations once for every nonzero in $\vect{A}$, the overall computational complexity of FusedMM is $O(d*\nnz)$.
For memory estimations, we assume 8 byte indices and single precision values. 
Thus, $\vect{X}$ needs $4md$ bytes, $\vect{Z}$ needs $4md$ bytes, $\vect{Y}$ needs $4nd$ bytes, and $\vect{A}$ needs $12\nnz$ bytes to store. 
Thus the total memory requirement of FusedMM is $8md{+}4nd{+}12\nnz$ bytes.
If the fused kernel is not used, one needs to store the intermediate matrix $\vect{H}$ that may require $12\nnz*d$ bytes.
Thus, FusedMM needs asymptotically less memory than separate SDDMM and SpMM operations. 

To estimate the peak performance of FusedMM, we compute the arithmetic intensity (AI) which is the ratio of floating point operations to the bytes moved. Our SIMD vectorized FusedMM implementation optimizes the number of memory accesses for $\vect{X}$ and $\vect{Z}$ to their optimal values to $md$. The sparse matrix $\vect{A}$ is streamed only once ($\nnz$). However, the number of accesses for $\vect{Y}$ can be $\nnz*d$ (assuming no spatial or temporal locality when accessing $\vect{Y}$). 
Assuming $\delta$ to be the average degree of the graph, AI is bounded as follows:
\begin{equation}
\label{eqn:bestAI}
    \text{AI} > \frac{2dm\delta + 2dm\delta} {12m\delta + 8md + 4dm\delta}  = \frac{\delta } { 3\frac{\delta}{d} + 2 + \delta}
\end{equation}
where we used $\nnz{=}m\delta$ and considered both addition and multiplications as floating point operations. 
Clearly, AI depends on both average degree $\delta$ and feature dimension $d$. 
To see the relative influence of $\delta$ and $d$, we rearrange Eqn.~\ref{eqn:bestAI} as $(3/d + 2/\delta + 1)^{-1}$.
Hence for a typical embedding dimension of  $d=128$, AI is mostly determined by the sparsity of the graph. 
For denser graphs ($\delta\gg2$) with $d\gg3$, AI approaches to its best value of 1. 
The worst AI of 1/6 is obtained when the graph is very sparse with $\delta=1$ and $d=1$.
Therefore, we expect better performance for denser graphs.
Nevertheless, FusedMM is still memory bound for all reasonable values of $d$ and $\delta$. 
Hence, its peak performance is bounded by the memory bandwidth.

\section{Experiments}
\vspace{-2pt}
\subsection{Experimental Setup}
\vspace{-3pt}
\textbf{Overview of experiments.} We perform experiments 
with kernels needed by the force-directed graph layout based on the Fruchterman-Reingold (FR) model (Fig.~\ref{fig:motivation}(a)), the graph embedding (Fig.~\ref{fig:motivation}(b)), and GCN (Fig.~\ref{fig:motivation}(c)) algorithms. 
We primarily focus on the kernel timing on three different architectures. We also show end-to-end training and accuracy to assess the quality.


\textbf{Experimental platforms.} We conduct all of our experiments on three different servers with Intel, AMD, and ARM processors as described in Table \ref{tab:hardware}. 
We have implemented FusedMM in C/C++ programming language with OpenMP multi-threading 
and SIMD intrinsic support. 
Our code generator can automatically generate intrinsic codes for different architectures. 
In our experiments, we only considered single-precision values for $\vect{X}$, $\vect{Y}$, and $\vect{Z}$.
However, our code generator can generate efficient codes for double precision values as well.
Except in the scalability experiment, we use all available cores in each processor. 
For all of our experiments, we measure the time for 10 iterations and report the average time. 

\begin{table}[!t]
\centering
\caption{Hardware configurations of our experiments.}
\vspace{-3pt}
\label{tab:hardware}
\begin{tabular}{c|c|c|c|c|}
\cline{2-5}
\textbf{}                                          & \textbf{Property} & \textbf{\begin{tabular}[c]{@{}c@{}}Intel\\ Skylake  8160\end{tabular}}   & \textbf{\begin{tabular}[c]{@{}c@{}}AMD\\ EPYC 7551\end{tabular}} & \textbf{\begin{tabular}[c]{@{}c@{}}ARM ThunderX\\ CN8890\end{tabular}} \\ \hline
\multicolumn{1}{|c|}{\multirow{4}{*}{\parbox[t]{2mm}{\multirow{1}{*}{\rotatebox[origin=c]{90}{Core}}}}}        & Clock             & 2.10 GHz                                                           & 2 GHz                                                            & 1.9 GHz                                                         \\[0.18ex] 
\multicolumn{1}{|c|}{}                             & L1 cache          & 32KB   &                              32KB                        & 32KB                                                              \\[0.18ex] 
\multicolumn{1}{|c|}{}                             & L2 cache          & 1MB   &                              512KB                        & $\times$ \\[0.18ex] 
\multicolumn{1}{|c|}{}                             & LLC          & 32MB                                                                & 8MB                                                            & 16MB                                                                \\[0.18ex] \hline
\multicolumn{1}{|c|}{\multirow{3}{*}{\parbox[t]{2mm}{\multirow{1}{*}{\rotatebox[origin=c]{90}{\vspace{-2cm} Node}}}}} & Sockets           & 2                                                                  & 2                                                                & 1                                                               \\ [0.18ex]
\multicolumn{1}{|c|}{}                             & Cores/soc.        & 24                                                                 & 32                                                    & 48                                                              \\ [0.18ex]
\multicolumn{1}{|c|}{}                             & Memory            & 256GB                                                              & 128GB                                                            & 64GB                                                            \\ \hline
\multicolumn{1}{|c|}{\multirow{2}{*}{\parbox[t]{2mm}{\multirow{1}{*}{\rotatebox[origin=c]{90}{Env.}}}}}  & Compiler          & gcc 10.1.0                                                         & gcc 5.4.0                                                        & gcc 7.5.0                                                       \\ 
\multicolumn{1}{|c|}{}                             & Flags             & \begin{tabular}[c]{@{}c@{}}O3, mavx512f, \\ mavx512dq\end{tabular} & \begin{tabular}[c]{@{}c@{}}O3, mavx,\\ mfma\end{tabular}         & \begin{tabular}[c]{@{}c@{}}O3, asimd,\\ armv8-a\end{tabular}        \\ \hline
\end{tabular}
\vspace{-0.3cm}
\end{table}

\textbf{Baselines.}
We use DGL (version 0.5.2) on top of PyTorch (version 1.5.1) as a baseline to compare most of our results. DGL supports a C++ backend where users can implement their own C++ functions and integrate to DGL. 
In addition, it has native multi-threaded SDDMM and SpMM implementations in C++. 
DGL also provides a python message passing API for application developments.
These features give us an opportunity to integrate FusedMM with DGL, develop various high-level algorithms, and then make fair comparisons between FusedMM and DGL kernels. For both FusedMM and DGL, we measure runtime from the python interface. 
We also compare the SpMM specialization of FusedMM (i.e., the third row in Table~\ref{tab:fusedMM-operation}) with MKL (version 2019.5.281). 
In general, we term FusedMM to represent the SIMD vectorized implementation of our kernel except in Table~\ref{tab:intelkerneltime}. 

\textbf{Datasets.}
Table \ref{tab:datasets} shows a diverse set of graphs used in our experiments. It includes graphs having various number of vertices and edges, high average degree, low average degree, power-law property, etc. Some of them (e.g., Cora and Pubmed) are widely used to benchmark graph embedding and GNN algorithms. We also generate several RMAT graphs by PaRMAT \cite{wsvr} to assess the parameter sensitivity of FusedMM.

\begin{table}[!ht]
\centering
\caption{Graph datasets used in our experiments. Graphs are available at http://networkrepository.com and https://sparse.tamu.edu/ 
}
\vspace{-3pt}
\label{tab:datasets}
\begin{tabular}{c|c|c|c|c}
\hline
\textbf{Graphs} & \textbf{\#Vertices} & \textbf{\#Edges} & \textbf{Avg. Degree} & \textbf{Max. Degree} \\ \hline
Cora            & 2708                & 5278             & 3.90                 & 168                  \\ 
Harvard         & 15126               & 824617           & 109.03               & 1183                 \\ 
Pubmed          & 19717               & 44324            & 4.49                 & 171                  \\ 
Flickr          & 89250               & 449878           & 10.08                & 5425                 \\ 
Ogbprot.  &   132534      &   39561252    &   597   &   7750    \\ 
Amazon          & 334863              & 925872           & 5.59                 & 549                  \\ 
Youtube         & 1138499             & 2990443          & 5.25                 & 28754                \\ 
Orkut           & 3072441             & 117185083        & 76.28                &        33313              \\ \hline
\end{tabular}
\vspace{-0.4cm}
\end{table}


\begin{table*}
\centering
\caption{Kernel time (in sec.) for Graph Embedding, FR model, and GCN on Inter   server. A `$\times$' sign represents memory allocation error (i.e., out of memory). The columns 32, 64, 128, 256, and 512 represent the dimensions ($d$). FusedMM and FusedMMopt represent the general implementation and SIMD vectorized implementation of our proposed kernel, respectively. Speedup is computed for FusedMMopt over DGL.}
\vspace{-3pt}
\label{tab:intelkerneltime}

\arrayrulecolor[rgb]{0.8,0.8,0.8}
\begin{tabular}{c!{\color{black}\vrule}c!{\color{black}\vrule}m{0.45cm}|m{0.45cm}|m{0.45cm}|m{0.55cm}|m{0.65cm}!{\color{black}\vrule}m{0.55cm}|m{0.55cm}|m{0.55cm}|m{0.55cm}|m{0.45cm}!{\color{black}\vrule}c|m{0.55cm}|m{0.55cm}|m{0.55cm}|m{0.5cm}} 
\arrayrulecolor{black}\hline
\multicolumn{1}{l|}{}           & \multicolumn{1}{l|}{}             & \multicolumn{5}{c!{\color{black}\vrule}}{\textbf{Graph Embedding}}                                                                                                                                                          & \multicolumn{5}{c!{\color{black}\vrule}}{\textbf{FR model}}                                                                                                                                                               & \multicolumn{5}{c}{\textbf{GCN}}                                                                                                                                                                                         \\ 
\cline{3-17}
\multirow{2}{*}{\textbf{Graphs}} & \multirow{2}{*}{\textbf{Methods}} & \multicolumn{5}{c!{\color{black}\vrule}}{Dimensions (d)}                                                                                                                                                                    & \multicolumn{5}{c!{\color{black}\vrule}}{Dimensions (d)}                                                                                                                                                                  & \multicolumn{5}{c}{Dimensions (d)}                                                                                                                                                                                       \\ 
                                 &                                   & 32                                        & 64                                        & 128                                       & 256                                        & 512                                        & 32                                         & 64                                         & 128                                        & 256                                        & 512                                   & 32                                        & 64                                         & 128                                        & 256                                        & 512                                    \\ 
\hline
\multirow{4}{*}{Ogbprot.}      & DGL                               & 0.766                                     & 1.394                                     & 3.275                                     & 8.077                                      & 18.236                                     & 2.547                                      & 4.915                                      & 11.115                                     & 23.320                                     & $\times$                                     & 0.859                                     & 1.644                                      & 3.71                                       & 8.681                                      & $\times$                                      \\ 
\arrayrulecolor[rgb]{0.8,0.8,0.8}\cline{2-17}
                                 & FusedMM                           & 0.506                                     & 0.859                                     & 1.648                                     & 3.016                                      & 5.703                                      & 0.510                                      & 0.892                                      & 1.737                                      & 3.124                                      & 5.921                                 & 0.343                                     & 0.498                                      & 0.872                                      & 1.442                                      & 2.579                                  \\ 
\cline{2-17}
                                 & FusedMMopt                        & 0.226                                     & 0.247                                     & 0.345                                     & 0.775                                      & 1.358                                      & 0.222                                      & 0.249                                      & 0.323                                      & 0.730                                      & 1.409                                 & 0.114                                     & 0.122                                      & 0.166                                      & 0.449                                      & 0.74                                   \\ 

                                 & Speedup                           & {\cellcolor[rgb]{0.851,0.918,0.827}}3.385 & {\cellcolor[rgb]{0.851,0.918,0.827}}5.655 & {\cellcolor[rgb]{0.851,0.918,0.827}}9.488 & {\cellcolor[rgb]{0.851,0.918,0.827}}10.428 & {\cellcolor[rgb]{0.851,0.918,0.827}}13.433 & {\cellcolor[rgb]{0.851,0.918,0.827}}11.487 & {\cellcolor[rgb]{0.851,0.918,0.827}}19.737 & {\cellcolor[rgb]{0.851,0.918,0.827}}34.389 & {\cellcolor[rgb]{0.851,0.918,0.827}}31.947 & {\cellcolor[rgb]{0.851,0.918,0.827}}- & {\cellcolor[rgb]{0.851,0.918,0.827}}7.535 & {\cellcolor[rgb]{0.851,0.918,0.827}}13.475 & {\cellcolor[rgb]{0.851,0.918,0.827}}22.349 & {\cellcolor[rgb]{0.851,0.918,0.827}}19.334 & {\cellcolor[rgb]{0.851,0.918,0.827}}-  \\ 
\arrayrulecolor{black}\hline
\multirow{4}{*}{Youtube}         & DGL                               & 0.112                                     & 0.234                                     & 0.493                                     & 1.121                                      & 2.628                                      & 0.192                                      & 0.340                                      & 0.638                                      & 1.335                                      & 3.007                                 & 0.091                                     & 0.168                                      & 0.338                                      & 0.765                                      & 1.798                                  \\ 
\arrayrulecolor[rgb]{0.8,0.8,0.8}\cline{2-7}\cline{9-17}
                                 & FusedMM                           & 0.033                                     & 0.055                                     & 0.090                                     & 0.161                                      & 0.296                                      & 0.032                                      & 0.049                                      & 0.099                                      & 0.165                                      & 0.306                                 & 0.026                                     & 0.037                                      & 0.061                                      & 0.119                                      & 0.226                                  \\ 
\cline{2-17}
                                 & FusedMMopt                        & 0.026                                     & 0.032                                     & 0.058                                     & 0.123                                      & 0.226                                      & 0.024                                      & 0.033                                      & 0.057                                      & 0.121                                      & 0.231                                 & 0.019                                     & 0.035                                      & 0.061                                      & 0.106                                      & 0.164                                  \\ 

                                 & Speedup                           & {\cellcolor[rgb]{0.851,0.918,0.827}}4.255 & {\cellcolor[rgb]{0.851,0.918,0.827}}7.258 & {\cellcolor[rgb]{0.851,0.918,0.827}}8.463 & {\cellcolor[rgb]{0.851,0.918,0.827}}9.080  & {\cellcolor[rgb]{0.851,0.918,0.827}}11.647 & {\cellcolor[rgb]{0.851,0.918,0.827}}7.899  & {\cellcolor[rgb]{0.851,0.918,0.827}}10.290 & {\cellcolor[rgb]{0.851,0.918,0.827}}11.174 & {\cellcolor[rgb]{0.851,0.918,0.827}}11.007 & {\cellcolor[rgb]{0.851,0.918,0.827}}13.045 & {\cellcolor[rgb]{0.851,0.918,0.827}}4.789 & {\cellcolor[rgb]{0.851,0.918,0.827}}4.800  & {\cellcolor[rgb]{0.851,0.918,0.827}}5.541  & {\cellcolor[rgb]{0.851,0.918,0.827}}7.217  & {\cellcolor[rgb]{0.851,0.918,0.827}}10.963  \\ 
\arrayrulecolor{black}\hline
\multirow{4}{*}{Orkut}           & DGL                               & 1.760                                     & 3.336                                     & 6.851                                     & 15.734                                     & 34.014                                     & 4.044                                      & 7.682                                      & 14.098                                     & $\times$                                         & $\times$                                    & 1.045                                     & 1.922                                      & 3.993                                      & 8.137                                      & $\times$                                      \\ 
\arrayrulecolor[rgb]{0.8,0.8,0.8}\cline{2-2}\cline{4-7}\cline{9-17}
                                 & FusedMM                           & 0.969                                     & 1.601                                     & 3.247                                     & 5.441                                      & 9.665                                      & 0.993                                      & 1.662                                      & 3.352                                      & 5.975                                      & 9.758                                 & 0.746                                     & 1.076                                      & 2.077                                      & 3.71                                       & 6.083                                  \\ 
\cline{2-7}\cline{9-17}
                                 & FusedMMopt                        & 0.346                                     & 0.523                                     & 0.951                                     & 3.117                                      & 4.961                                      & 0.327                                      & 0.506                                      & 0.978                                      & 3.036                                      & 5.369                                 & 0.15                                      & 0.241                                      & 0.451                                      & 1.462                                      & 2.543                                  \\ 

                                 & Speedup                           & {\cellcolor[rgb]{0.851,0.918,0.827}}5.089 & {\cellcolor[rgb]{0.851,0.918,0.827}}6.381 & {\cellcolor[rgb]{0.851,0.918,0.827}}7.202 & {\cellcolor[rgb]{0.851,0.918,0.827}}5.048  & {\cellcolor[rgb]{0.851,0.918,0.827}}6.856  & {\cellcolor[rgb]{0.851,0.918,0.827}}12.372 & {\cellcolor[rgb]{0.851,0.918,0.827}}15.192 & {\cellcolor[rgb]{0.851,0.918,0.827}}14.414 & {\cellcolor[rgb]{0.851,0.918,0.827}}-      & {\cellcolor[rgb]{0.851,0.918,0.827}}- & {\cellcolor[rgb]{0.851,0.918,0.827}}6.967 & {\cellcolor[rgb]{0.851,0.918,0.827}}7.975  & {\cellcolor[rgb]{0.851,0.918,0.827}}8.854  & {\cellcolor[rgb]{0.851,0.918,0.827}}5.566  & {\cellcolor[rgb]{0.851,0.918,0.827}}-  \\
\arrayrulecolor{black}\hline
\end{tabular}
\vspace{-0.4cm}
\end{table*}

\subsection{Kernel time performance}
\vspace{-3pt}
We integrate FusedMM into DGL and implement graph embedding, FR model, and GCN algorithms discussed in Fig.~\ref{fig:motivation} and Table~\ref{tab:fusedMM-operation}. We also implement these algorithms using the SDDMM and SpMM kernels of DGL. Then, we run
FusedMM-based algorithms and DGL's kernel based algorithms from the python interface and measure the runtimes of FusedMM and DGL kernels (excluding IO and prepossessing).

\textbf{Performance on Intel Server.} We report the kernel time of FusedMM and DGL in Table \ref{tab:intelkerneltime} for graph embedding, FR graph layout model, and GCN. Here, FusedMMopt represents the SIMD vectorized implementation of FusedMM. Due to space restriction, we choose Ogbprot., Youtube, and Orkut graphs as representatives of high average degree, low average degree, and bigger size graph, respectively. To analyze the efficiency of register blocking, we show the results for various dimensions (i.e., $d$) of $\vect{X}$. Our key findings are listed below.



(1) {\em FusedMM even without any optimization runs up to $9.8\times$ faster than DGL}.The speedups of unoptimized FusedMM relative to DGL are: best $9.8\times$ for $d=512$ with FR model on Youtube graph; worst $1.4\times$ for $d=32$ with GCN on Orkut graph; average $4.2\times$ over all graphs and dimensions.
These results clearly demonstrate the benefit of a fused kernel that does not store intermediate matrices.

(2) {\em The optimized FusedMM kernel gives us up to $5.4\times$ speedups over unoptimized FusedMM.}
The speedups of FusedMMopt relative to unoptimized FusedMM are: best $5.4\times$ for $d{=}128$ with FR model using Ogbprot.; worst $1.0\times$ for $d{=}128$ with GCN using Youtube; average $2.7\times$ over all graphs and dimensions.
These results clearly demonstrate the effectiveness of vectorization and autotuned intrinsic operations.

(3) {\em DGL may go out of memory when high-dimensional messages are used.}
The FR graph layout algorithm generates $d$-dimensional messages on each edge. This may require prohibitive memory to store the intermediate matrix $\vect{H}$.
For example, Table~\ref{tab:intelkerneltime} shows that DGL goes out of memory on Orkut for 256 and 512 dimensions. 
As FusedMM does not store $\vect{H}$, it is more robust with respect to feature dimensions. 
For the same reason, FusedMMopt achieves its best speedup over DGL when used in the FR model.

(4) {\em FusedMM performs better on denser graphs.} Table~\ref{tab:intelkerneltime} shows that FusedMM performs better on Ogbprot., the most dense graph in our test suite.  
This is expected because denser graph can amortize memory latency costs and usually have higher arithmetic intensities. 

(5) {\em FusedMM performs better for higher dimensions.}
Table~\ref{tab:intelkerneltime} clearly shows that FusedMMopt becomes more effective at higher dimensions (relative to DGL). 
This is due to FusedMM's register blocking strategy that efficiently utilizes available registers with $d$-dimensional feature vectors.

Overall, FusedMMopt comprehensively outperforms DGL by a significant margin on all graphs for all dimensions. 
It is possible that the speedup of FusedMMopt drops a little after the dimension $128$ (especially in GCN) because we may observe some register spilling at higher dimensions.  

\begin{figure}
    \centering
    \includegraphics[width=0.7\linewidth]{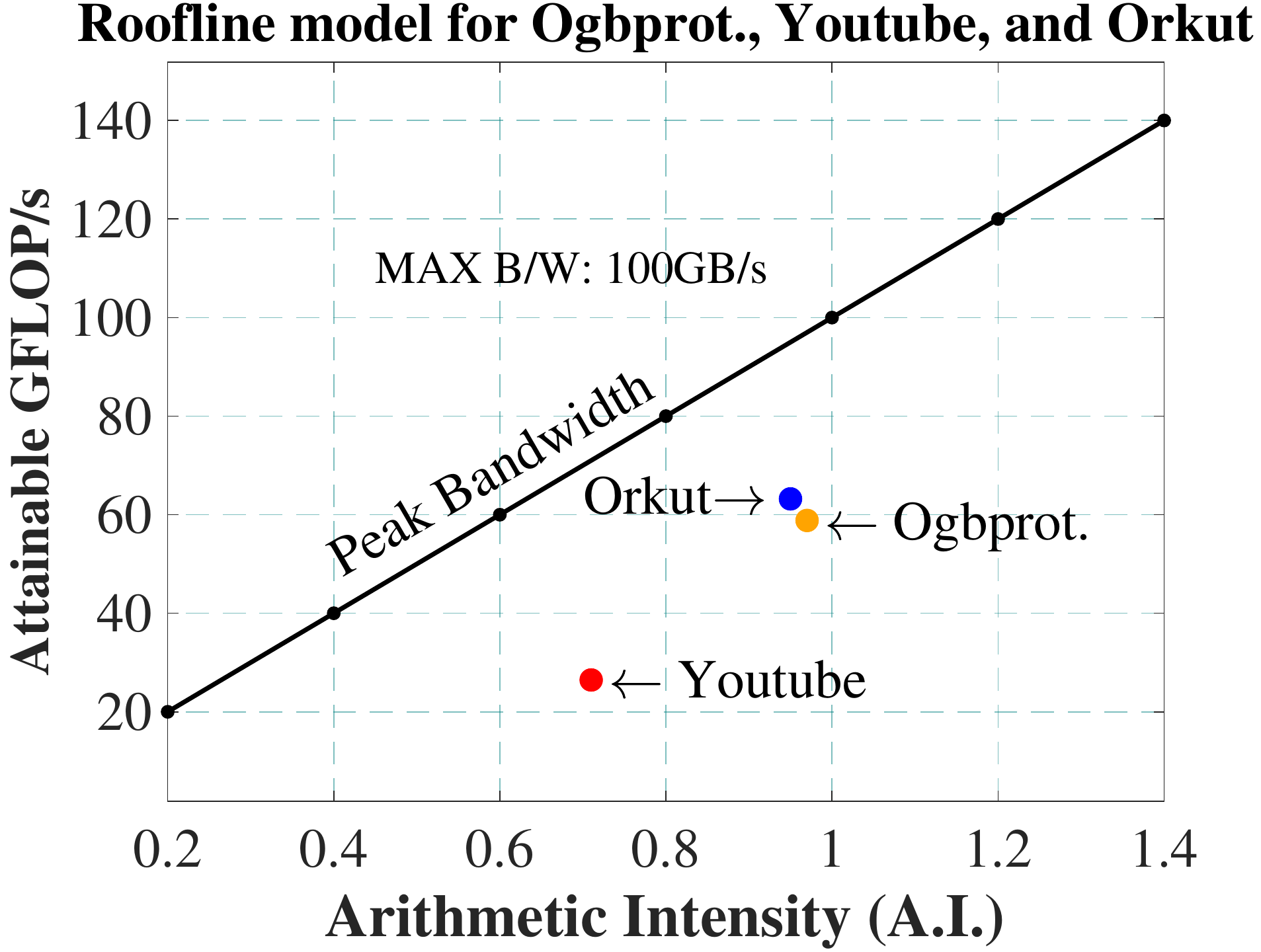}
    \vspace{-3pt}
    \caption{Roofline model of FusedMM for Ogbprot., Youtube, and Orkut graphs on Intel server for graph embedding. The STREAM bandwidth on this server is 100 GB/s. We set $d$ to $128$.}
    \label{fig:roofline}
    \vspace{-3pt}
\end{figure}

\textbf{Roofline analysis of FusedMM.} Based on Eq.~\ref{eqn:bestAI}, we show a roofline model~\cite{williams2009roofline} of the graph embedding task on Intel server in Fig. \ref{fig:roofline}. We observe that FusedMM achieves $63.21$ GFLOP/s for the Orkut graph with an AI of 0.95. For Orkut, the best possible performance is $95.27$ GFLOP/s according to the Roofline model. 
Similarly for other graphs, the observed performance is reasonably good, but  they fall a little short of the best possible performance.
This gap between the observed and attainable performance comes from the overheads associated with  python function calls.
When we directly called FusedMM in a C++ code, the observed performance is very close to the attainable performance. 
We still report the performance observed from the python interface because this performance is realized by end users. 




\textbf{Comparison with Intel MKL.} 
Since MKL \cite{mkl} does not have an SDDMM operation, we only compare SpMM-based GCN ($3^{rd}$ row of Table \ref{tab:fusedMM-operation}) with MKL's SpMM function.
For this experiment, we measure the kernel time of SpMM from the C++ interface. 
We measure both inspection and execution time for MKL.
Table \ref{tab:spmmkerneltime} shows that the SpMM specialization of FusedMM performs comparably to MKL's SpMM implementation.
Thus, despite being a multipurpose kernel, FusedMM can match the best performing specializations of SpMM.
\begin{table}[!htb]
\centering
\caption{Kernel time (in sec.) of SpMM on Intel server for various dimensions. Best value is marked in \textbf{bold}.}
\vspace{-3pt}
\label{tab:spmmkerneltime}
\begin{tabular}{c|c|m{0.38cm}m{0.5cm}c|m{0.38cm}m{0.38cm}c}
\hline
\multirow{2}{*}{\textbf{Graphs}} & \multirow{2}{*}{\textbf{Method}} & \multicolumn{3}{c|}{\textbf{Single Thread}}        & \multicolumn{3}{c}{\textbf{48 Threads (2 soc.)}} \\
                                 &                                  & 64             & 128             & 256             & 64              & 128            & 256            \\ \hline
\multirow{2}{*}{Ogbprot.}        & MKL                              & 1.017          & 2.310           & 5.318           & 0.034           & 0.094          & \textbf{0.264} \\
                                 & FusedMM                          & \textbf{0.951} & \textbf{1.990}  & \textbf{4.125}  & \textbf{0.031}  & \textbf{0.075} & 0.336          \\ \hline
\multirow{2}{*}{Youtube}         & MKL                              & 0.142          & 0.310           & 0.606           & \textbf{0.012}  & 0.031          & \textbf{0.071} \\
                                 & FusedMM                          & \textbf{0.132} & \textbf{0.261}  & \textbf{0.524}  & 0.015  & \textbf{0.028} & 0.082          \\ \hline
\multirow{2}{*}{Orkut}           & MKL                              & 6.336          & 14.356          & 29.348          & \textbf{0.380}  & 0.852          & \textbf{1.961} \\
                                 & FusedMM                          & \textbf{5.876} & \textbf{11.897} & \textbf{23.292} & 0.389  & \textbf{0.828} & 2.775          \\ \hline
\end{tabular}
\vspace{-3pt}
\end{table}

\begin{figure*}[!t]
    \centering
    \fbox{\includegraphics[width=0.26\linewidth,height=3.3cm]{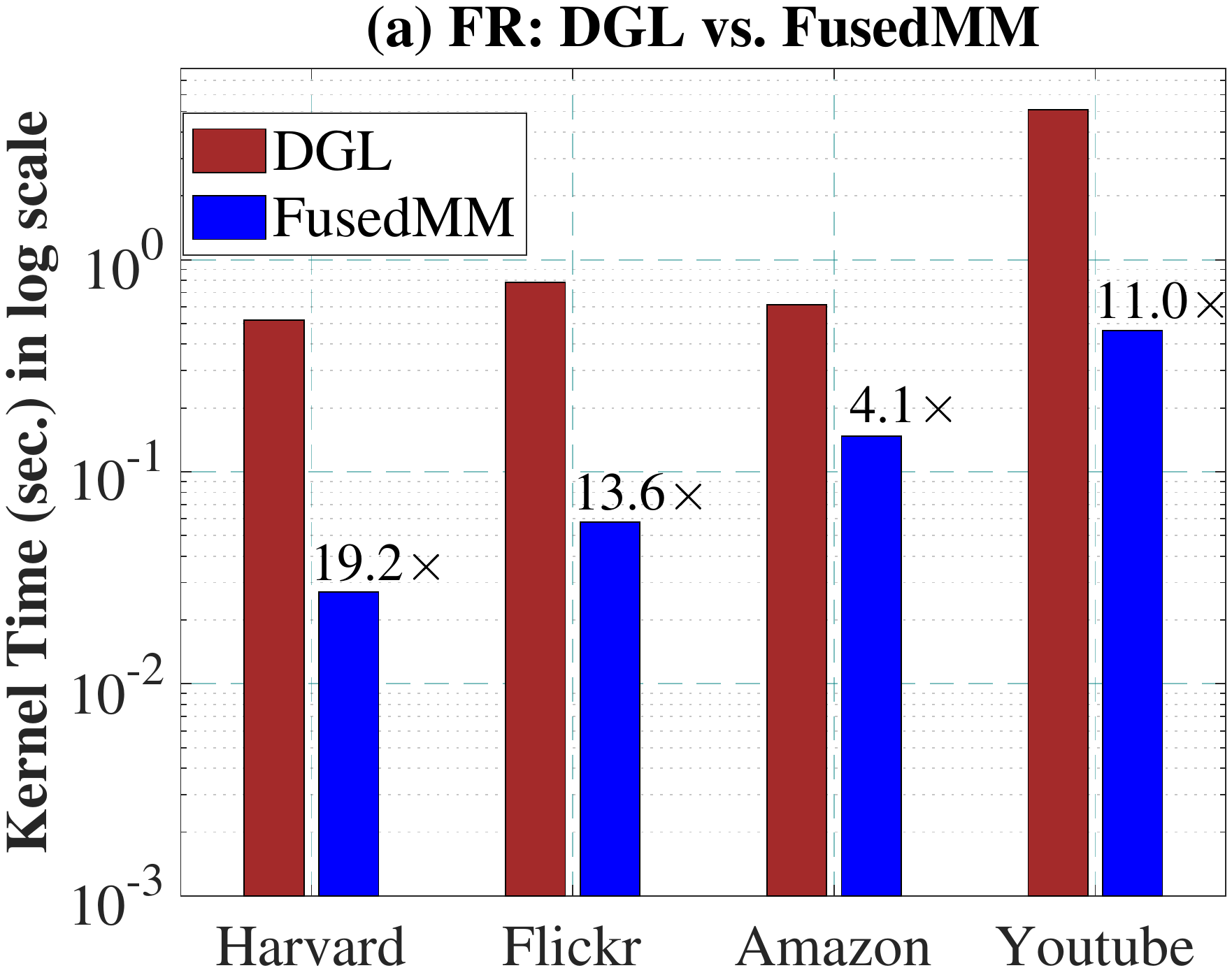}} ~~
    \fbox{\includegraphics[width=0.26\linewidth,height=3.3cm]{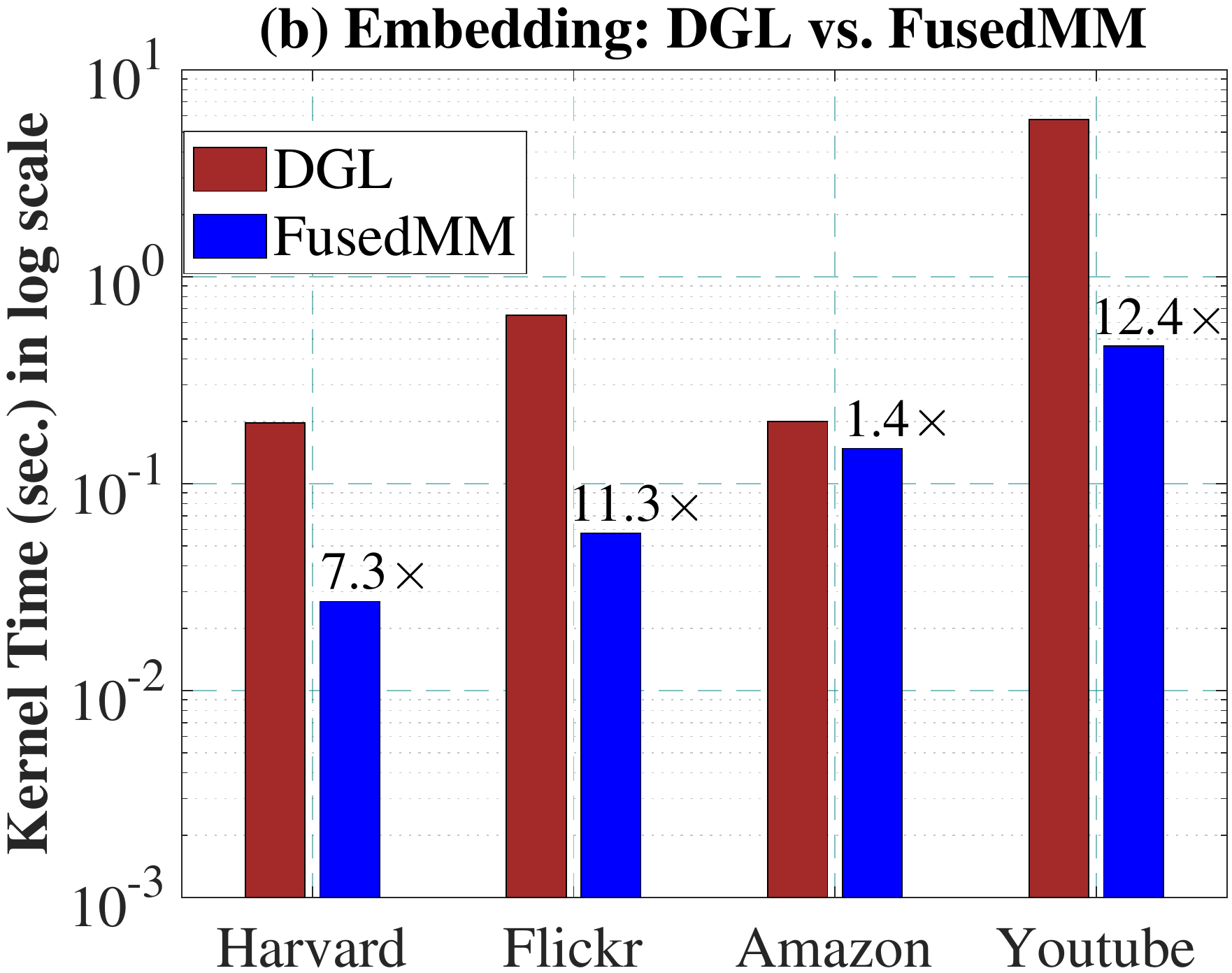}} ~~
    \fbox{\includegraphics[width=0.26\linewidth,height=3.3cm]{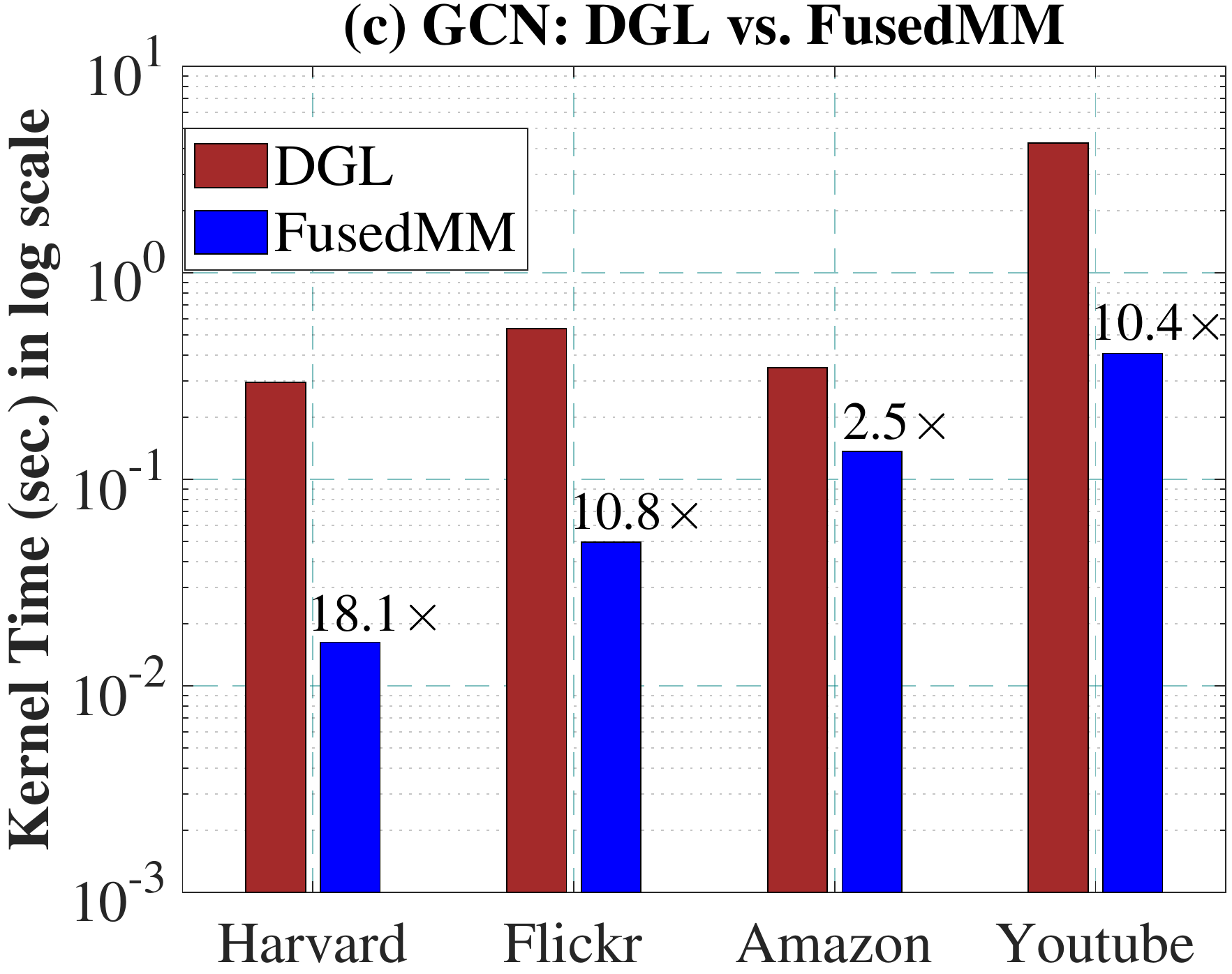}}
    \caption{Kernel time of FusedMM and DGL on ARM server using various benchmark graphs ($d=128$) for (a) FR model, (b) Graph Embedding, and (C) GCN. In all the figures, the speedup of FusedMM over DGL is shown above its representative (blue colored) bars.}
    \label{fig:abcarm}
\end{figure*}

\begin{figure}[!t]
    \centering
    \includegraphics[width=0.48\linewidth,height=3.2cm]{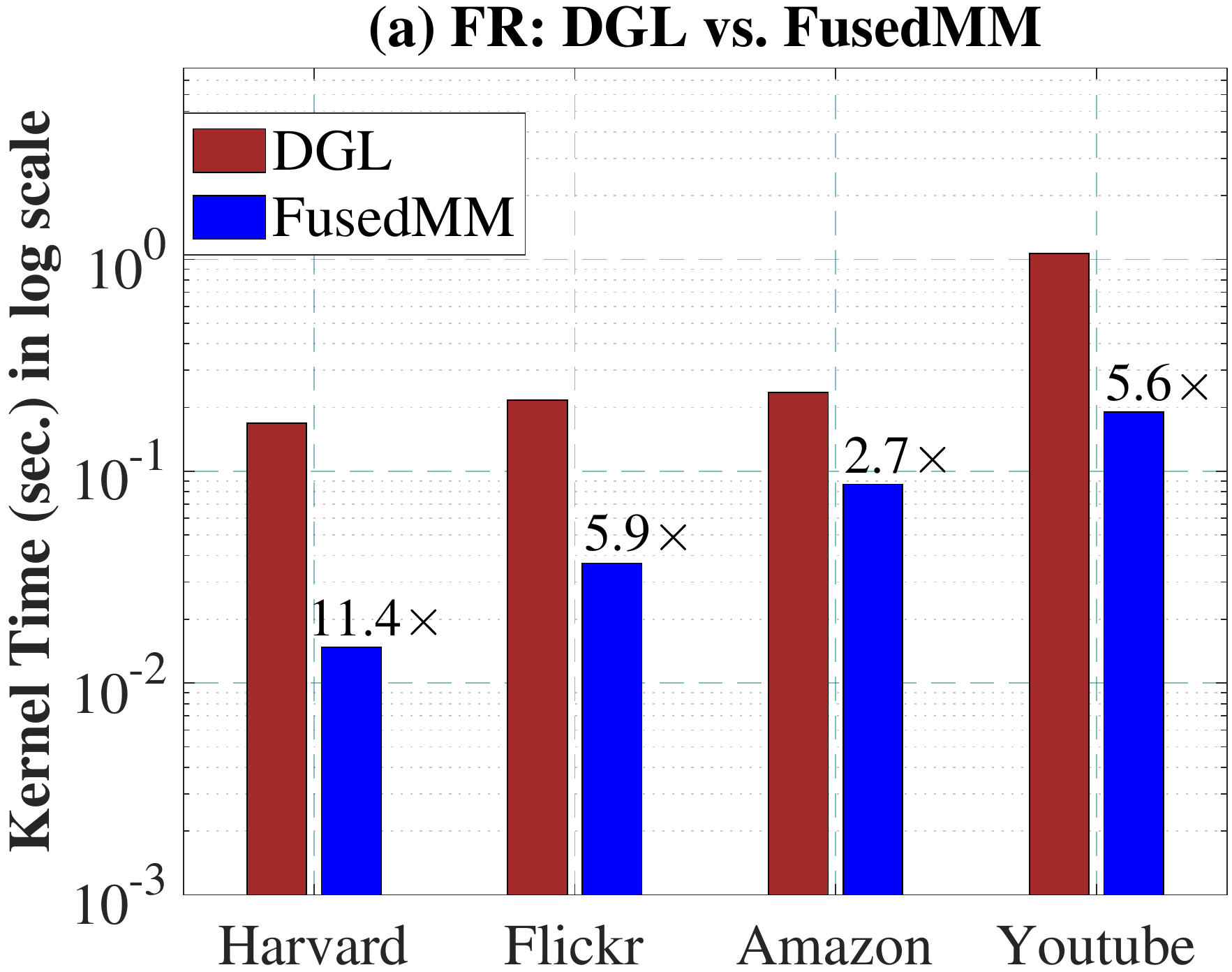}
    \includegraphics[width=0.48\linewidth,height=3.2cm]{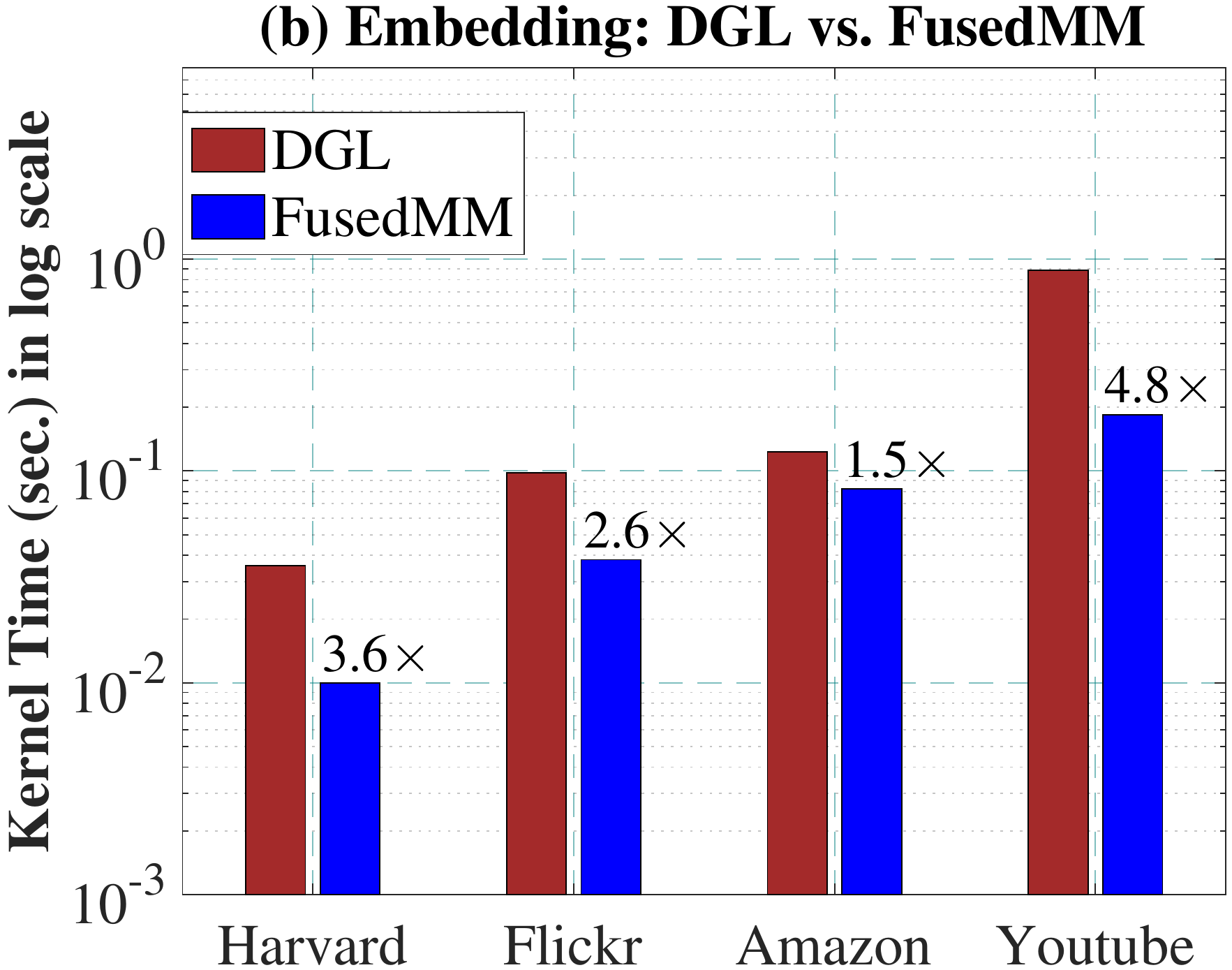}
    \caption{Kernel time of FusedMM and DGL on AMD server using various benchmark graphs (here, $d=128$) for (a) FR model, and (b) Graph Embedding. In all the figures, the speedup of FusedMM with respect to DGL is shown above its representative (blue colored) bars.}
    \label{fig:abcamd}
\end{figure}

\textbf{Performance on ARM ThunderX and AMD EPYC.} We conduct similar experiments on ARM and AMD servers and report the results in Figs. \ref{fig:abcarm} and \ref{fig:abcamd}. For all these results, we use $d{=}128$. Due to memory limitations, we could not generate results for Ogbprot. and Orkut. In Fig. \ref{fig:abcarm} (a), we observe that FusedMM is up to $19.2\times$ faster than DGL. 
As with the Intel processor, the observed performance originates from the fused design and effective register blocking of SIMD vectorization. As our ARM server has no L2 cache, FusedMM gets the full advantage of register blocking. Fig. \ref{fig:abcamd} shows that 
FusedMM achieves up to $11.4\times$ speedup over DGL. 
Notably, optimized codes for ARM and AMD processors were autotuned using our code generator. 
Hence, application developers do not need to write optimized codes for different architectures.

\vspace{-2pt}
\subsection{Sensitivity Analysis}
\begin{figure}
    \centering
    \includegraphics[width=0.49\linewidth,height=3.7cm]{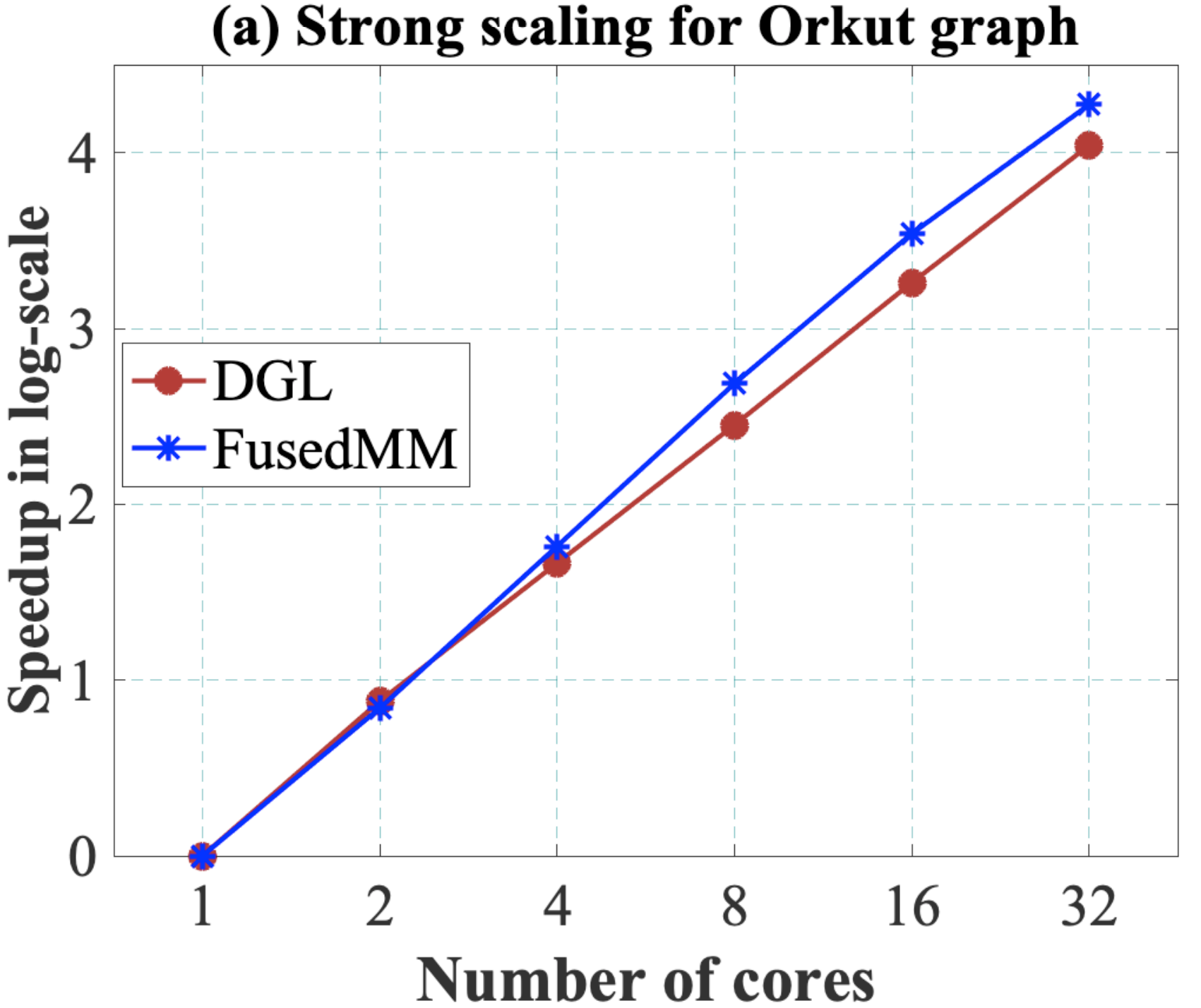}
    \includegraphics[width=0.49\linewidth,height=3.7cm]{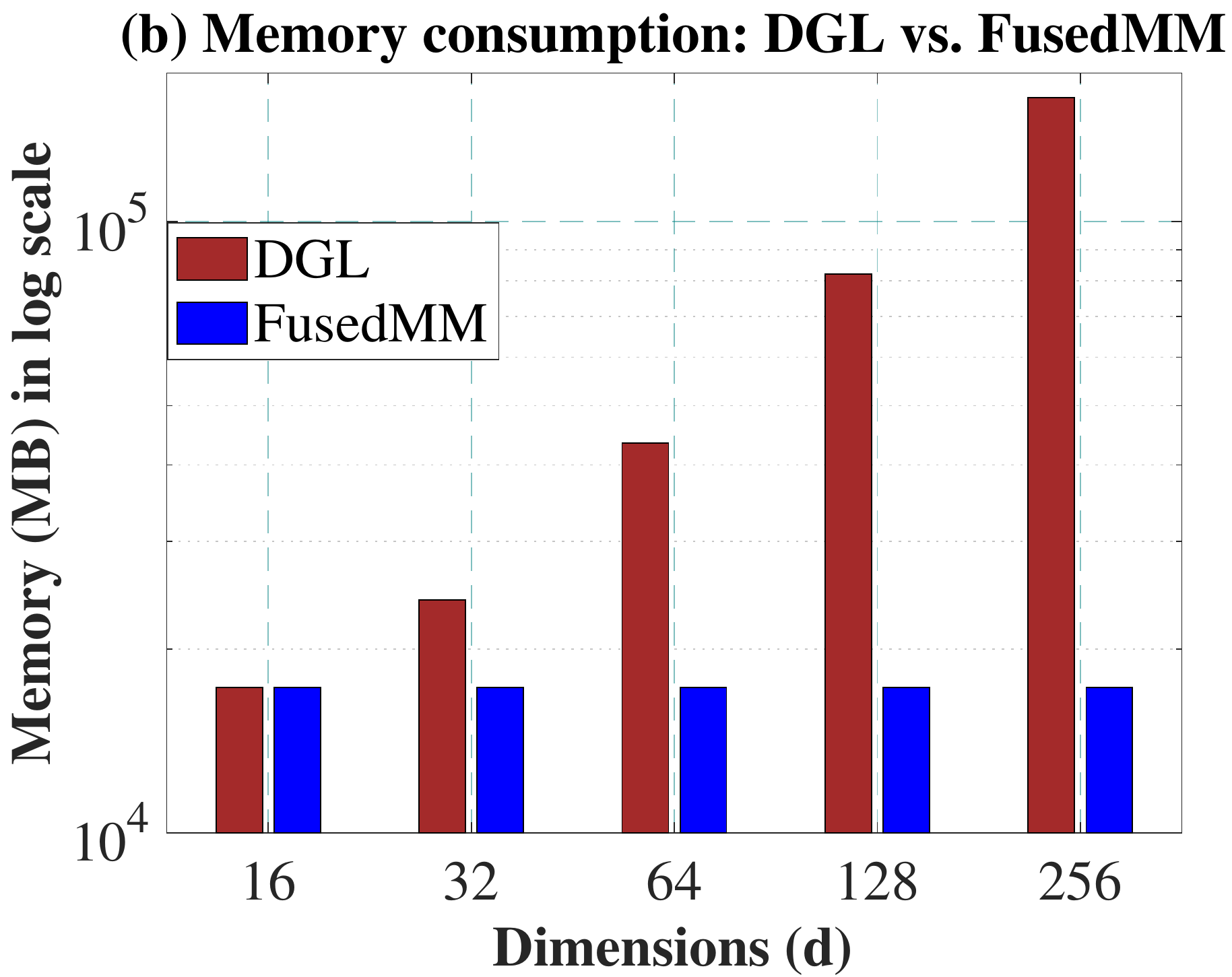}
    \vspace{-15pt}
    \caption{(a) Strong scaling of FusedMM and DGL for Graph Embedding using Orkut graph with respect to their sequential execution (here, $d=256$). (b) Memory consumption in megabytes in the FR model of DGL and FusedMM for Ogbprot.}
    \label{fig:scalibilitymem}
    \vspace{-8pt}
\end{figure}

\vspace{-3pt}
\textbf{Scalability.} Fig. \ref{fig:scalibilitymem} shows the scalability of FusedMM and DGL for graph embedding. 
We perform this experiment on the Intel server for Orkut graph with $d{=}256$.
We observe that FusedMM on 32 cores is ${\sim}20\times$ faster than its sequential runtime. 
DGL's kernels also scale well achieving up to $16\times$ speedup, but runs slower than FusedMM for all thread counts. 


\textbf{Memory consumption.} When SDDMM and SpMM operates in tandem in DGL, we need to store intermediate results in $\vect{H}$ (Eq. \ref{eqn:sddmmout}). This can consume a significant amount of memory when an application (e.g.,  FR graph layout) generates a sparse-tensor as depicted in Fig.~\ref{fig:messagepassing}. 
Fig.~\ref{fig:scalibilitymem}(b) shows that DGL's memory requirement grows linearly with $d$ for the FR model while the memory consumption of FusedMM remains stable. 
This gives FusedMM a clear advantage over unfused kernels in DGL for tasks that require high-dimensional messages.  

\begin{figure}
    \centering
    \includegraphics[width=0.48\linewidth, height=3.7cm]{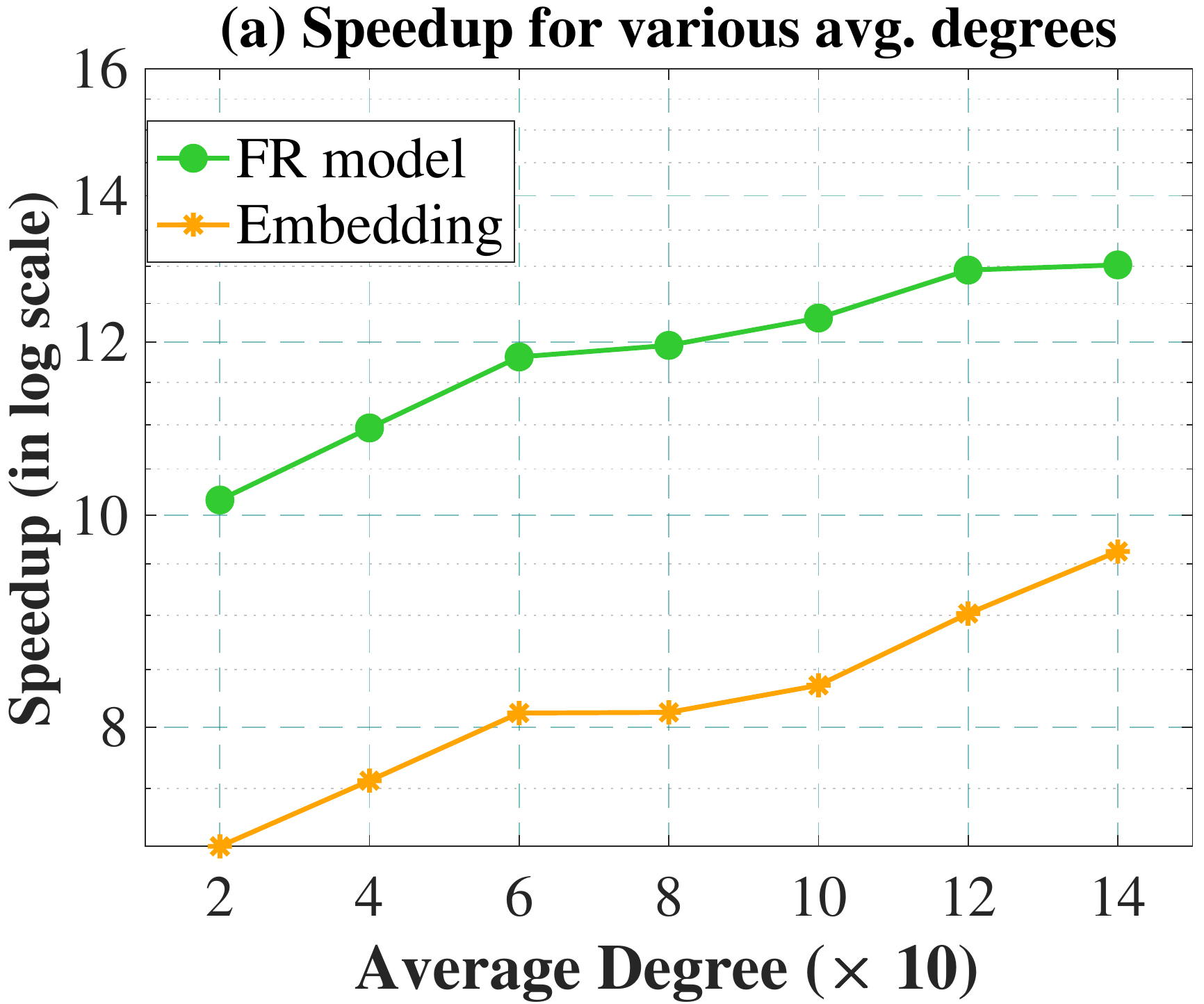}
    \includegraphics[width=0.48\linewidth, height=3.7cm]{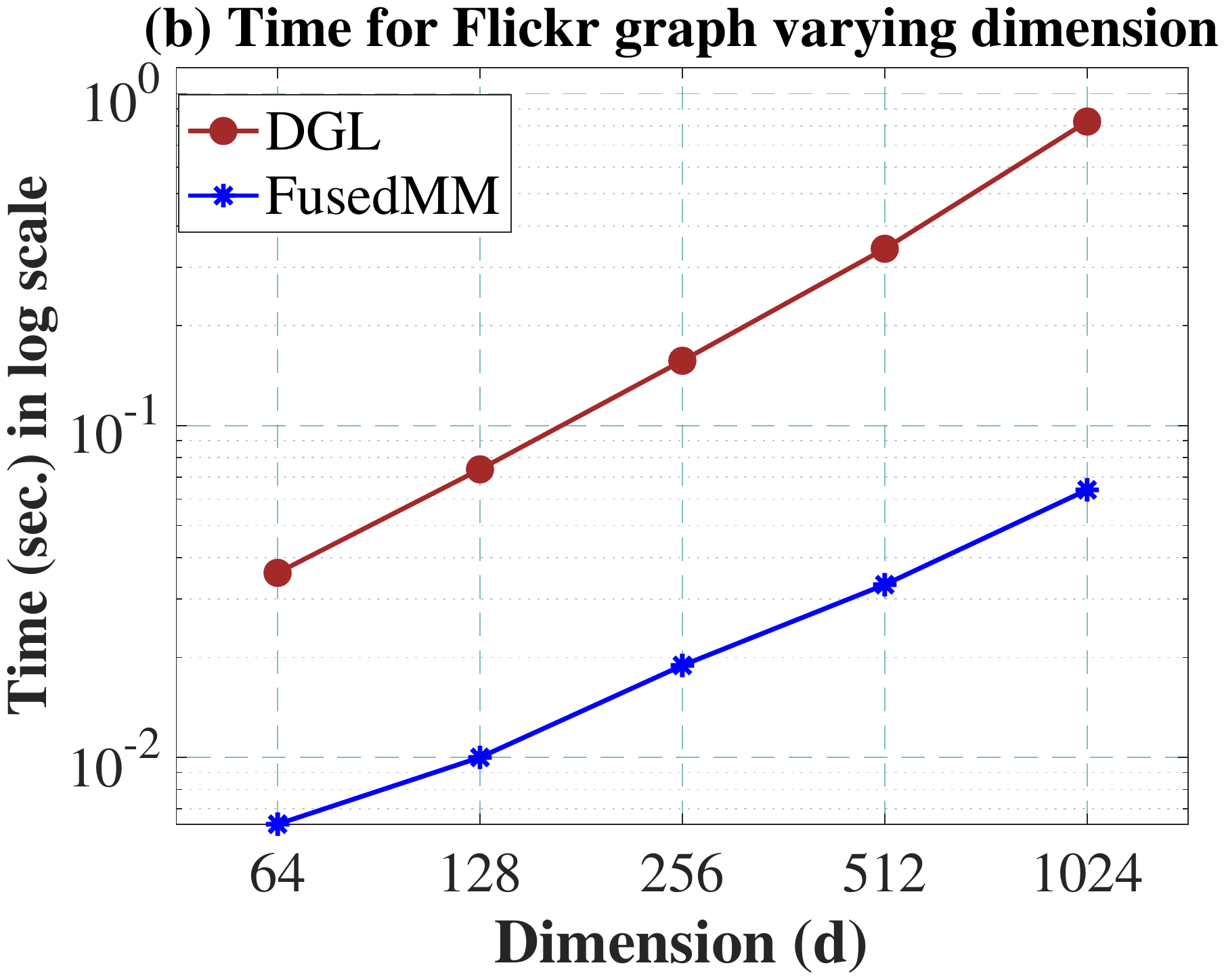}
    \vspace{-4pt}
    \caption{(a) Speedup of FusedMM over DGL for various RMAT graphs with 100K vertices and various average degrees. (b) Kernel time of Flickr graph for graph embedding varying the size of $d$.}
    \label{fig:averagedegree}
    \vspace{-4pt}
\end{figure}

\textbf{Parameter sensitivity.} 
We study the performance of FusedMM by changing average degrees of graphs and feature dimensions. 
Fig. \ref{fig:averagedegree}(a) shows the speedup of FusedMM over DGL for various average degrees of RMAT graphs with 100K vertices. The initial graph has one million edges and we increase this by a factor of 2. We observe that the speedup of FusedMM increases with the increase in the average degree. These results are consistent for both the FR model and graph embedding. In Fig. \ref{fig:averagedegree} (b), we show the kernel time of FusedMM and DGL for Flickr varying the dimension. We observe that both kernels show similar sensitivity to $d$. FusedMM is significantly faster than DGL for all values of $d$, and their performance gap widens as $d$ increases.

\begin{table}
\centering
\caption{Graph Embedding application time per-epoch for different methods ($d=128$, and batch size is 256). 
}
\vspace{-3pt}
\label{tab:end2end}
\begin{tabular}{c|c|c|c} 
\hline
                            Graphs      &       Method                                     & Total Time (Sec.)           & Speedup  \\ 
\hline
\multirow{3}{*}{Cora}             & PyTorch                           & 0.342                  & 48.9$\times$         \\
                                  & DGL                               & 0.177~                  & 25.3$\times$         \\
                                  & FusedMM                           & \textbf{0.007}  & 1.0$\times$           \\ 
\hline
\multirow{3}{*}{Pubmed}           & PyTorch                           & 2.590                  & 45.4$\times$         \\
                                  & DGL                               & 1.415                    & 28.3$\times$         \\
                                  & FusedMM                           & \textbf{0.057}  & 1.0$\times$            \\
\hline
\end{tabular}
\vspace{-5pt}
\end{table}


\vspace{-2pt}
\subsection{End-to-End training} 
\vspace{-3pt}
We developed three implementations of the  Force2Vec~\cite{force2vec} graph embedding algorithm to perform end-to-end training. We implement Force2Vec using standard kernels in PyTorch, SDDMM and SpMM kernels in DGL, and using FusedMM. 
We set embedding dimension to 128, batch size to 256, and the number of epochs to 800. Due to the high mini-batch processing time of DGL, we use Cora and Pubmed for this experiment. These two graphs are widely used to benchmark graph embedding and graph neural network methods \cite{kipf2016semi,huang2020ge}. 
When the end-to-end training is considered, Table \ref{tab:end2end} shows that FusedMM is up to $25\times$ and $45\times$ faster than DGL and PyTorch, respectively.
FusedMM performs significantly better than DGL because FusedMM can directly take a scaling operations (SOP in Table~\ref{tab:fusedMM-operation}).
By contrast, DGL needs to generate the intermediate results to perform the scaling.  


\textbf{Accuracy:} Finally, we assess the quality of the embedding generated by the Force2Vec algorithm implemented using FusedMM. 
As FusedMM does not alter the actual computations performed, we do not expect any performance loss compared to the original implementation of Force2Vec.
Indeed the original Force2Vec and FusedMM-based Force2Vec both achieve the same F1-micro scores of 0.78 and 0.79 when performing node classifications on Cora and Pubmed datasets.


\section{Related Work}
Graphs and sparse matrices are fundamentally related, and their 
duality~\cite{kepner2011graph} has been exploited in many graph algorithms~\cite{azad2018hipmcl,trianglegabb15,shun2013ligra} and libraries~\cite{davis2019algorithm, bulucc2011combinatorial,sundaram2015graphmat,kepner2016mathematical,malewicz2010pregel}.
Over the last few years, graph ML algorithms have been increasingly using linear algebra kernels to capture various message passing operations.   
For example, the original GCN implementation from Kipf and Welling~\cite{kipf2016semi} used SpMM to capture the graph convolution operation. 
SpMM-like operations were also used in high-performance graph layout~\cite{rahman2020batchlayout} and embedding~\cite{force2vec} algorithms. 

Recently, several specialized frameworks were developed to make the processing of GNN workloads easier and faster.
Among them PyG~\cite{fey2019pyg} and DGL~\cite{wang2019dgl} provides message-passing APIs to develop high-level applications.
However, they use linear algebra kernels in the back end for performance. 
DGL explicitly calls SDDMM and SpMM implementations for message generation and aggregation operations. Hence, FusedMM can directly substitute linear algebra kernels used by DGL. Note that the fusion of SDDMM and SpMM kernels has also been used to compute word mover's distance \cite{tithi2020efficient}.

When standard SpMM and SDDMM kernels are used in GNN, PyG  and  DGL  call vendor-provided libraries  (e.g.,  MKL \cite{mkl} and  cuSPARSE \cite{cusparse}) because these libraries provide highly optimized implementations for sparse kernels. 
However, DGL also provides general implementations of SpMM and SDDMM kernels to capture complex graph convolutions and attention mechanisms  used in GNNs \cite{velivckovic2017graph,thekumparampil2018attention,hamilton2017inductive,kipf2016semi,chen2018fastgcn}. 
Recently, Huang et al.~\cite{huang2020ge} developed a general-purpose SpMM algorithm for GPUs.
When integrated with DGL and PyG, their GE-SpMM kernel can expedite the computations of GCN and pooling based GNNs such as GraphSAGE~\cite{hamilton2017inductive}. 
FeatGraph~\cite{hu2020featgraph} provides efficient implementations of SDDMM and SpMM for both CPUs and GPUs. 
In that sense, FusedMM developed in this paper is a generalization of FeatGraph.

%


\vspace{-1pt}
\section{Discussions and Conclusions}
\vspace{-2pt}
We present a flexible linear algebra kernel called FusedMM that captures the core computations of most graph embedding and graph neural network algorithms.
Conventionally, graph learning libraries such as PyG and DGL rely on at least two matrix operations for edge-wise message generations and vertex-wise message aggregations.
FusedMM substitutes them with just one general-purpose operation with the support of uses-defined functions. 
Our results are unexpectedly positive because FusedMM even without any optimization runs significantly faster than equivalent kernels used in DGL. 
This clearly demonstrates the value of reducing memory traffic using fused operations in graph machine learning.
Even though graph algorithms are harder to vectorize due to irregular computations, we were able to make  FusedMM up to $5\times$ faster using our automatically tuned vectorized operations.
Thus, this paper brings in the philosophy of 
Automatically Tuned Linear Algebra Software (ATLAS)~\cite{atlas_sc98} in sparse computations, which was an unexplored territory in graph analytics. 
Fused kernels and autotuning approaches together make high-level graph learning algorithms at least an order of magnitude faster than unfused and untuned kernels.

 While this paper only considers CPU implementations, the vectorization and autotuning techniques developed for FusedMM are easily applicable to GPUs as well.
 On GPUs, we will employ threads (SIMT) in place of lanes of registers (SIMD) such that each thread in a block will compute an element of the row of the dense matrix. 
 Thus, FusedMM provides a general-purpose technique to accelerate graph analysis and graph machine learning.


\section*{Acknowledgement}

We would like to thank anonymous reviewers for their feedback. Funding for this work was provided by the Indiana University Grand Challenge Precision Health Initiative.

\bibliographystyle{IEEEtran}
\bibliography{references}

\end{document}